\documentclass[11pt]{article}
\usepackage[skins]{tcolorbox}
\tcbuselibrary{listings, breakable}
\usepackage{multirow}
\usepackage{booktabs}
\usepackage{makecell}
\usepackage{amsmath,amssymb}
\usepackage{siunitx}
\usepackage{enumitem}
\usepackage{algorithm}
\usepackage{algpseudocode}

\usepackage[table]{xcolor} 
\usepackage[preprint]{acl}
\usepackage{times}
\usepackage{latexsym}
\usepackage[T1]{fontenc}
\usepackage[utf8]{inputenc}
\usepackage{microtype}
\usepackage{inconsolata}
\usepackage{graphicx}

\usepackage{tabularx}
\usepackage{colortbl}
\newcolumntype{Y}{>{\centering\arraybackslash}X}
\newcommand{\method}{PACE}

\title{Enhancing Multimodal RLVR with Pairwise Token Contrast and Adaptive Trajectory Confidence}

\author{
  Yiran Zhang$^{1,2}$ \quad Hanbo Huang$^{1}$ \quad Jielan Li$^{2}$ 
  \quad Zhen Zhao$^{2}$\thanks{Corresponding author} 
  \quad Shiyu Liang$^{1}$\footnotemark[1]
  \quad Hongxia Hao$^{2}$\footnotemark[1] \\
  $^{1}$Shanghai Jiao Tong University \quad $^{2}$ Shanghai Artificial Intelligence Laboratory \\
  \texttt{\{yrzhangz, lsy18602808513\}@sjtu.edu.cn} \quad \texttt{\{haohongxia, zhaozhen\}@pjlab.org.cn}
}

\begin{document}
\maketitle
\begin{abstract}
Reinforcement learning with verifiable rewards (RLVR) improves multimodal reasoning, but group-relative policy optimization broadcasts a single sequence-level advantage to every token of a response, although errors usually localize to a few decisive steps. Recent self-distilled variants refine credit with a teacher--student likelihood gap. However, this signal cannot tell whether it relies on the image or on a language shortcut, and is applied uniformly even when the teacher is uninformative. To address this, we propose \textbf{P}airwise token contrast and \textbf{A}daptive trajectory \textbf{C}onfidenc\textbf{E} (\textbf{\method{}}). Pairwise token contrast scores each response under four teacher views, crossing a correct or incorrect answer hint with the image present or removed, and compares the two hints for every token, which cancels shifts shared by both hints and reveals whether the token's support relies on the image. Adaptive trajectory confidence measures how well the teacher predicts each response given the correct hint and the image, normalizes this score within the batch, and uses it to set how far the token contrast redistributes credit within that response. \method{} achieves the best average accuracy on three backbones and ranks first on all evaluated multimodal reasoning benchmarks with Qwen3-VL-8B-Instruct. Experiments show that both components matter: replacing pairwise token contrast with a single-view signal underperforms our method, while adaptive trajectory confidence adds further improvement.

\end{abstract}

\begin{figure*}[t]
  \centering
  \includegraphics[width=\textwidth]{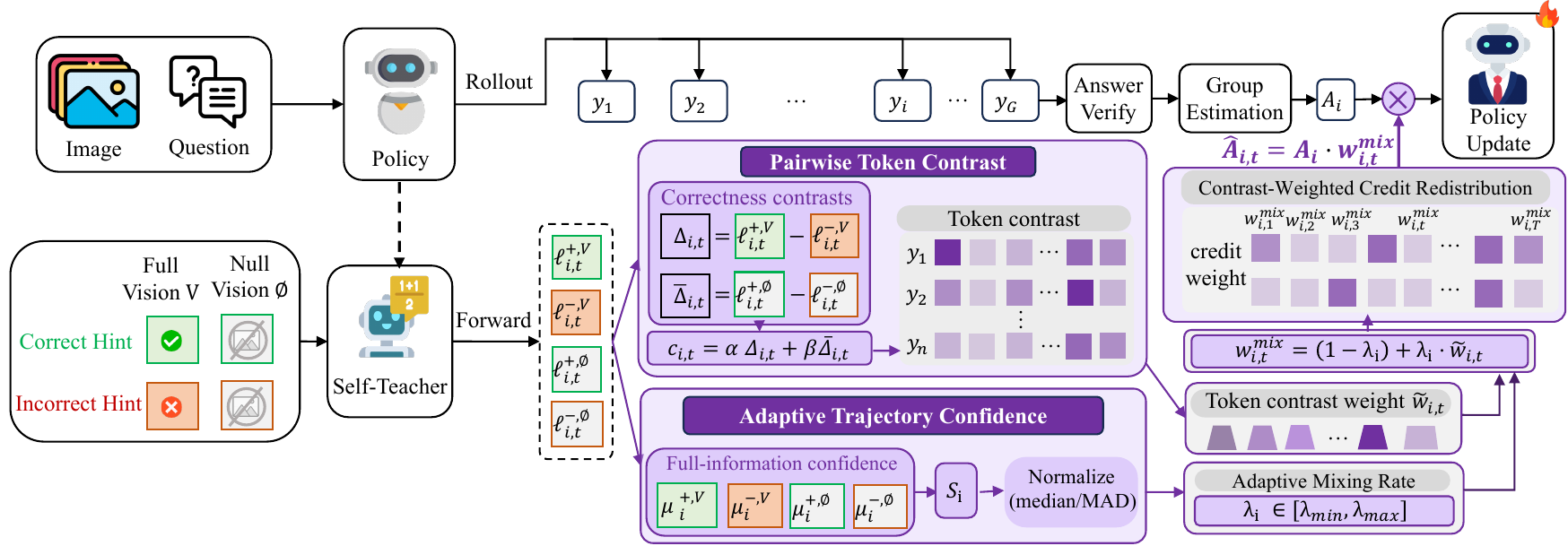}
  \caption{Overview of \method{}. A periodically refreshed teacher scores each sampled response under four counterfactual views (correct/incorrect hints $\times$ full/null vision). Pairwise token contrast yields the token contrast $c_{i,t}$; adaptive trajectory confidence uses the teacher's batch-normalized confidence on each response to set the mixing rate $\lambda_i$, i.e., how far $c_{i,t}$ redistributes credit among the tokens of each response.}
  \label{fig:overview}
  \vspace{-0.4cm}
\end{figure*}

\section{Introduction}
\label{sec:introduction}
RLVR has emerged as a powerful approach for improving reasoning in large language models~\citep{guo2025deepseek}. A rule-based verifier scores sampled responses, and group-relative policy optimization methods such as GRPO~\citep{shao2024deepseekmath} convert these verifier rewards into learning signals without requiring a learned reward model. This paradigm extends naturally to vision-language models, improving visual mathematics and multimodal reasoning~\citep{huang2026vision,meng2025mm}.

Self-distilled RLVR refines the sequence-level advantage of GRPO at the token level, as in RLSD~\citep{yang2026self} and RLCSD~\citep{pan2026rlcsd}. A periodically synchronized copy of the policy scores each sampled response under a privileged context, typically the ground-truth answer, and the resulting teacher--student likelihood gap reweights individual tokens. This lets credit concentrate on the few steps where long multimodal reasoning chains usually go wrong, such as a misread chart label or a single arithmetic mistake.

Although these approaches provide dense token-level supervision, they face two unaddressed challenges in multimodal reasoning. On the one hand, they derive this supervision from a contrast along a single modality. 
RLCSD varies only the textual answer hint, while visually grounded signals~\citep{wang2026perception,huang2026spotlight} vary only the image. 
On the other hand, they apply the same fixed weighting strategy to every rollout trajectory, while the teacher is confident on some trajectories and uninformative on others.
Together, these two challenges aggravate the drift of the teacher--student gap. The signal comes from a single view and is applied with equal strength to every response. Any shift in the teacher's or the student's likelihood during training passes directly into the token weights.

To address these limitations, we propose \method{}, a method for enhancing multimodal RLVR with two components: pairwise token contrast and adaptive trajectory confidence, as shown in Figure~\ref{fig:overview}. \emph{Pairwise token contrast} scores each sampled response under four teacher views that cross correct vs.\ incorrect answer hints with full vs.\ null vision. Comparing the two hints under full vision and under null vision gives two correctness contrasts per token, which we combine into a single token contrast. The null-vision contrast exposes whether the correctness signal persists without the image, and because the two views in each contrast differ only in the hint, offsets shared by both hints, including drift in the teacher's overall confidence, cancel out. \emph{Adaptive trajectory confidence} measures how much better the teacher predicts each response under the full-information view than under the degraded views, normalizes this confidence within the batch, and uses it to set how far the token contrast redistributes credit among the tokens of each response. Our contributions are as follows:
\begin{itemize}[leftmargin=*,itemsep=0pt,topsep=2pt]
\item We propose pairwise token contrast, which removes the dependence on the student's likelihood, cancels confidence drift shared by both hints, and assigns credit without biasing it toward either the image or the text.
\item We propose adaptive trajectory confidence, which treats distinct trajectories differently by applying the token contrast more strongly where the teacher is confident and less where it is uninformative, remains stable as the teacher's confidence drifts during training, and preserves the sign and mean of every response's advantage.
\item Extensive experiments show that \method{} obtains the best average accuracy on all three main backbones across five multimodal reasoning benchmarks. We provide ablations showing that pairwise token contrast and adaptive trajectory confidence both contribute to the gains.
\end{itemize}

\section{Related Works}
\paragraph{RLVR for language and multimodal model reasoning.}
Reinforcement learning with verifiable outcome avoids a learned reward model when correctness can be checked with rules. GRPO estimates relative advantages from multiple responses to the same prompt without a separate critic~\citep{shao2024deepseekmath}, and multimodal variants apply this recipe to visual mathematics and general visual reasoning~\citep{huang2026vision,meng2025mm}. These methods generally broadcast a sequence-level advantage across response tokens. \method{} keeps the verifier and group advantage unchanged and assigns that credit at token level.

\paragraph{On-policy self-distillation and token-level credit assignment.}
On-policy distillation~\citep{agarwal2024policy} and its self-distilled variants evaluate responses sampled by the student itself~\citep{zhao2026self,yang2026self,hubotter2026reinforcement,li2026unifying,pan2026rlcsd}. Recent work makes the same broad move toward finer credit: VCSD uses image-content removal to form a visual contrast and distills a sharpened token distribution~\citep{liang2026visual}; GCPO contrasts positive and negative prompts and derives token advantages from predictive divergence~\citep{li2026guidance}; PEPO combines a hidden-state perception prior with token entropy~\citep{li2026rethinking}; and CEPO, the closest contemporaneous method, contrasts a correct-answer teacher with a wrong-answer teacher built from an already-rejected rollout to flag decisive tokens on multimodal math reasoning~\citep{heakl2026cepo}. These methods differ from \method{} in both signal and objective. VCSD does not condition the contrast on correctness, GCPO does not factor out generic visual sensitivity with a full/null factorial design, PEPO uses perception--exploration statistics rather than answer-conditioned teacher contrasts, and CEPO's answer contrast holds the image fixed rather than crossing it with a full/null vision axis. \method{} uses answer-conditioned, vision-dependent token contrast for sign-preserving GRPO credit assignment, not as a standalone distillation target.

\paragraph{Visual grounding and reflective multimodal RL.}
PAPO, VPPO, and propagation-aware visual retention prioritize visual dependence or retention during long-chain reasoning~\citep{wang2026perception,huang2026spotlight,gong2026reflection}; reflection-centric systems instead add explicit rethink or re-attention stages~\citep{wang2026vl,wan2026srpo,yao2026mulberry,jian2025look,chu2025qwen}. Dependence alone is unsigned with respect to correctness, and reflection changes the sampled reasoning process. In contrast, \method{} evaluates the already sampled response under correct/incorrect hints and full/null vision, then uses the resulting token contrast for token-level credit assignment without changing inference.

\section{Preliminaries}
\label{sec:preliminaries}

\paragraph{Multimodal RLVR.}
Given a prompt $x=(q,V)$ with question $q$ and image $V$, the policy $\pi_\theta$ samples $G$ responses $\{y_i\}_{i=1}^{G}$, where $y_i=(y_{i,1},\dots,y_{i,T_i})$, and a rule-based verifier assigns each response a reward $r_i\in\{0,1\}$. GRPO~\citep{shao2024deepseekmath} normalizes the rewards within the group into $A_i=(r_i-\operatorname{mean}(\{r_j\}))/\operatorname{std}(\{r_j\})$ and maximizes the clipped surrogate
\begin{equation}
\mathcal{J}(\theta) = \mathbb{E}\!\left[ \tfrac{1}{G}\sum_{i,t} \tfrac{1}{T_i} \min\!\big( \rho_{i,t} \hat{A}_{i,t},\, \bar{\rho}_{i,t} \hat{A}_{i,t} \big) \right]
\label{eq:grpo}
\end{equation}
with $\hat{A}_{i,t}=A_i$, where $\rho_{i,t}$ is the token-level importance ratio to the rollout policy $\pi_{\theta_{\mathrm{old}}}$ and $\bar{\rho}_{i,t}$ its clipped value; the KL term is disabled. Every token of a response thus receives the same credit, although errors usually localize to a few decisive tokens. RLSD~\citep{yang2026self} instead sets $\hat{A}_{i,t}=A_i\cdot\operatorname{clip}(e^{\operatorname{sign}(A_i)\,g_{i,t}},1-\varepsilon_w,1+\varepsilon_w)$ with the detached gain
\begin{equation}
g_{i,t}=\log \bar\pi(y_{i,t}\,|\,x, h^{+}_i) - \log \pi_{\theta_{\mathrm{old}}}(y_{i,t}\,|\,x),
\label{eq:rlsd_gain}
\end{equation}
where the teacher $\bar\pi$ is a periodically synchronized copy of the policy and $h^{+}_i$ is the ground-truth answer given as a hint.

\paragraph{Limitations of self-distilled RLVR.}
The gain in Eq.~\eqref{eq:rlsd_gain} exhibits the two challenges raised in Section~\ref{sec:introduction}, which in turn make the gain drift. First, $g_{i,t}$ is measured under a single context that contains the image, so the gain cannot tell whether a token's support comes from the image or from a language-only pattern. Second, the same weighting is applied to every response, including those on which the teacher barely separates the privileged context from its alternatives. As a result, the single-view gain drifts: an offset in the teacher's log-probabilities, 
passes into $g_{i,t}$ directly, and $\log\pi_{\theta_{\mathrm{old}}}$ changes as the policy trains, so the gain drifts even when the token's dependence on the answer does not.

\paragraph{Problem formulation.}
We keep the verifier's judgment of each response and only redistribute its credit among tokens, i.e., we maximize Eq.~\eqref{eq:grpo} with $\hat{A}_{i,t}=A_i\,w_{i,t}$, where the detached weights satisfy $w_{i,t}>0$ and $\frac{1}{T_i}\sum_t w_{i,t}=1$. These constraints keep the sign of $A_i$ on every token and its mean over the response, and GRPO is the special case $w_{i,t}=1$. We further write $w^{\mathrm{mix}}_{i,t}=(1-\lambda_i)+\lambda_i\tilde{w}_{i,t}$, where $\tilde{w}_{i,t}$ is a unit-mean token weight derived from a token signal $c_{i,t}$ and $\lambda_i$ is a per-trajectory mixing rate. These issues then translate into requirements on the two factors: $c_{i,t}$ should be measured with and without the image, and should compare views that differ only in the answer hint so that shared offsets cancel; $\lambda_i$ should vary with the teacher's confidence on each response, measured relative to the current batch. Section~\ref{sec:methodology} instantiates $c_{i,t}$ as pairwise token contrast and $\lambda_i$ as adaptive trajectory confidence.

\section{Methodology}
\label{sec:methodology}

\begin{figure*}[t]
  \centering
  \includegraphics[width=\textwidth]{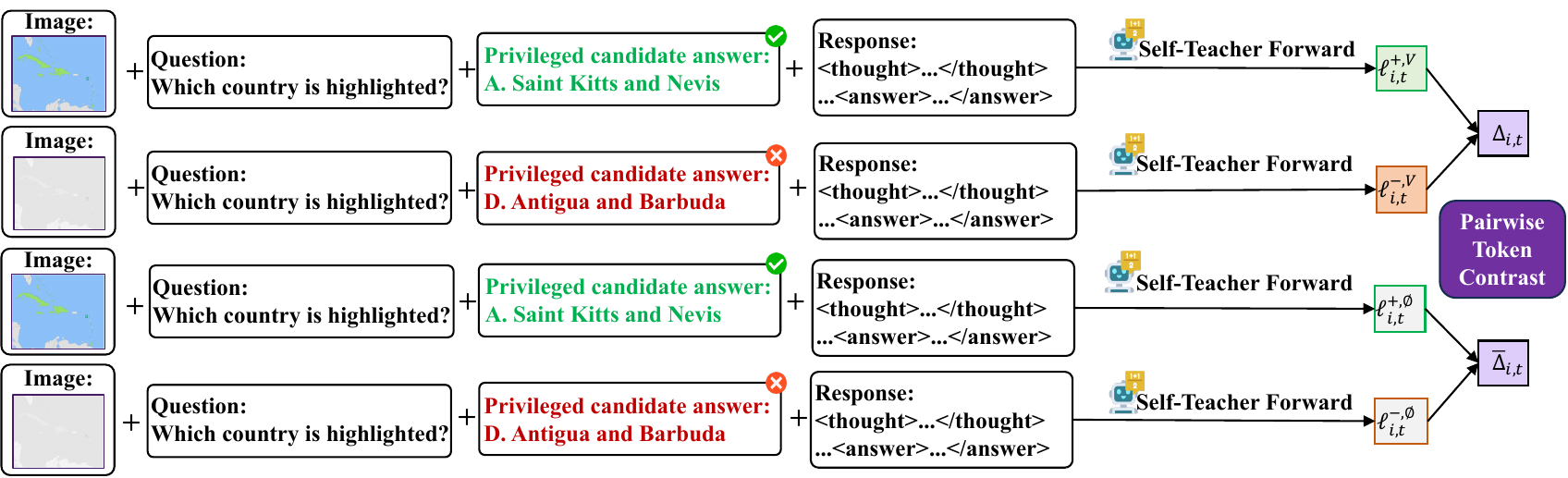}
  \caption{The four teacher views of pairwise token contrast.}
  \label{fig:pairwise_token_contrast}
  \vspace{-0.3cm}
\end{figure*}

\subsection{Counterfactual Teacher Views}
\label{sec:method_views}

Every sampled response $y_i$ is scored by the teacher $\bar\pi$ (a copy of the policy, hard-refreshed every 20 steps) under a $2\times2$ factorial design over two privileged factors: the \emph{answer hint} (correct vs.\ incorrect) and the \emph{visual input} (full vs.\ null), as the example illustrated in Figure~\ref{fig:pairwise_token_contrast}.

\paragraph{Candidate-answer hints.}
For each response we build a positive hint $h^{+}_i$ from the ground-truth answer, and collect the $K_i$ valid negative hints $\{h^{-}_{i,k}\}_{k=1}^{K_i}$ from \emph{incorrect} responses of sibling rollouts in the same GRPO group, deduplicated by their normalized final answers. Both branches share an identical template (\emph{``Privileged candidate answer: \dots''}), so the two contexts differ only in the answer content. Note that the positive branch contains a single ground-truth hint, while the negative branch aggregates $K_i$ incorrect hints; this asymmetry is deliberate, reflecting the one-vs-many structure of correctness contrasts. When $K_i=0$ (no incorrect siblings in the group, or all the same answer), the pairwise contrast cannot be formed, and the trajectory falls back to single-view RLSD weights or uniform weights (details in Sec.~\ref{sec:method_nta}).

\paragraph{Visual-null intervention.}
To remove visual evidence without touching the textual context, we replace the visual-token embeddings with a fixed zero vector; we write the intervened image as $\varnothing$. Teacher-forcing $y_i$ under these contexts yields the realized-token log-probabilities of the four views, where $q_i$ and $V_i$ are the question and image of the prompt that produced $y_i$. We write each view as $\ell^{a,b}_{i,t}$, where the first superscript $a\in\{+,-\}$ marks the correct or incorrect hint and the second $b\in\{V,\varnothing\}$ the full or null image. Under the positive hint and full vision,
\begin{equation}
\ell^{+,V}_{i,t} = \log \bar\pi\!\big(y_{i,t}\,|\,q_i, V_i, h^{+}_i, y_{i,<t}\big),
\label{eq:view_pos}
\end{equation}
and for the negative side we marginalize the $K_i$ valid sibling hints in probability space:
\begin{equation}
\ell^{-,V}_{i,t} = \log\!\Big[ \frac{1}{K_i}\sum_{k=1}^{K_i} \bar\pi\!\big(y_{i,t}\,|\,q_i, V_i, h^{-}_{i,k}, y_{i,<t}\big) \Big].
\label{eq:views}
\end{equation}
The visual-null counterparts $\ell^{+,\varnothing}_{i,t}$ and $\ell^{-,\varnothing}_{i,t}$ are defined identically with $V_i$ replaced by $\varnothing$. All teacher scores are detached, which only weight tokens and never contribute gradients.

\subsection{Pairwise Token Contrast}
\label{sec:method_contrast}

From the four views we form two \emph{correctness contrasts} of the same realized token:
\begin{equation}
\Delta_{i,t} = \ell^{+,V}_{i,t} - \ell^{-,V}_{i,t}, \qquad
\bar{\Delta}_{i,t} = \ell^{+,\varnothing}_{i,t} - \ell^{-,\varnothing}_{i,t}.
\label{eq:contrasts}
\end{equation}
$\Delta_{i,t}$ measures how strongly token $y_{i,t}$ is tied to the correct answer when full vision is available; $\bar{\Delta}_{i,t}$ repeats the measurement under the visual counterfactual, exposing how much of the correctness signal survives without looking at the image. The token contrast is a weighted combination of the two contrasts:
\begin{equation}
c_{i,t} = \alpha\,\Delta_{i,t} + \beta\,\bar{\Delta}_{i,t},
\label{eq:token_contrast}
\end{equation}
where $\alpha$ and $\beta$ are chosen such that $\alpha = 1+\beta$ to maintain a unit-weight anchor on $\Delta_{i,t}$ while varying the correction strength $\beta$. In practice, we set $(\alpha,\beta)=(1.5, 0.5)$ (Appendix~\ref{app:implementation_details}). With $\beta>0$, a token whose correctness preference holds with and without the image ($\Delta_{i,t}\approx\bar{\Delta}_{i,t}$) receives a token contrast $\approx2\Delta_{i,t}$, whereas a token preferring correctness only with the image ($\bar{\Delta}_{i,t}\approx0$) receives $1.5\Delta_{i,t}$; the null-vision contrast thus acts as a moderate second estimate 
rather than penalizing vision-independent tokens.

The sign of $c_{i,t}$ indicates whether the token supports the correct answer ($c_{i,t}>0$) or the incorrect ($c_{i,t}<0$). Whether this translates to reinforcement or suppression depends on both the token contrast \emph{and} the trajectory's verifier advantage: the token weight in Eq.~\eqref{eq:direction_aware_weight} is constructed as $\operatorname{sign}(A_i)\,c_{i,t}$. Thus, for positive trajectories ($A_i>0$), tokens with $c_{i,t}>0$ receive larger weights and are reinforced more, while for negative trajectories ($A_i<0$) the same tokens receive smaller weights and are penalized less; the penalty concentrates on tokens that favor the incorrect answer.

\paragraph{Why pairwise contrasts?}
A simpler alternative is to derive token weights without contrasting a hint pair, e.g.\ from the visual sensitivity $\ell^{+,V}_{i,t}-\ell^{+,\varnothing}_{i,t}$ (the teacher's confidence in a token with vs.\ without the image) or from the teacher--student gap of RLSD. Such single-view signals reflect how confident the teacher is under some context, not how much the token prefers the correct answer. The distinction matters because the teacher, refreshed from the improving policy, grows steadily more confident during training, e.g., under full vision. Such a drift shifts these signals even when the token's true dependence on the answer or the image is unchanged, misallocating credit (Figure~\ref{fig:training_signals}a in Appendix~\ref{app:training_signals}). Each contrast in Eq.~\eqref{eq:token_contrast} avoids this: its two sides share the same teacher, response, and visual context and differ only in the hint, so the shared drift cancels in the subtraction and only the token's preference for the correct answer remains, as detailed in Appendix~\ref{app:shared_offset}.

\subsection{Adaptive Trajectory Confidence}
\label{sec:method_nta}

The token contrast is not equally trustworthy on every trajectory: if the teacher fails to separate the privileged view from the degraded ones on response $y_i$, its per-token contrasts are mostly noise, and the safest choice is to reduce reliance on contrast weighting. We therefore score the teacher's \emph{full-information confidence} on each trajectory,
\begin{equation}
S_i = \mu^{+,V}_i - \tfrac{1}{3}\mu^{-,V}_i - \tfrac{1}{3}\mu^{+,\varnothing}_i - \tfrac{1}{3}\mu^{-,\varnothing}_i,
\label{eq:confidence}
\end{equation}
where $\mu^{+,V}_i$, $\mu^{-,V}_i$, $\mu^{+,\varnothing}_i$, $\mu^{-,\varnothing}_i$ are the within-trajectory means of the four views in Eqs.~\eqref{eq:view_pos}--\eqref{eq:views}. $S_i$ measures how much better the teacher predicts the response under the full-information view (correct hint and full vision) than under the three degraded views. It is large when the teacher achieves high confidence under full privileged information but lower certainty on any single degraded view, indicating that the response is well-resolved under the teacher's complete knowledge. Note that $S_i$ reflects overall teacher confidence rather than whether the response actually needs the image: it reduces reliance on the token contrast from trajectories where the teacher barely separates the views, but does not by itself test visual grounding, as detailed in  Appendix~\ref{app:nta_limitations}.

Because the absolute level of $S_i$ drifts as the policy moves, we normalize it robustly within each micro-batch over the set $\mathcal{V}$ of trajectories with a valid hint pair:
\begin{equation}
z_i = \operatorname{clip}\!\Big( \frac{S_i - \operatorname{med}_{\mathcal{V}}(S)}{\operatorname{MAD}_{\mathcal{V}}(S)},\, -\zeta,\, \zeta \Big),
\label{eq:normalize}
\end{equation}
where $\operatorname{med}_{\mathcal{V}}$ and $\operatorname{MAD}_{\mathcal{V}}$ are the median and median absolute deviation of $\{S_j\}_{j\in\mathcal{V}}$, and $\zeta{=}3$. The normalized confidence $z_i$ is then mapped to a per-trajectory mixing rate through a sigmoid:
\begin{equation}
\eta_i = \sigma(z_i),
\qquad
\lambda_i = \lambda_{\min} + \big(\lambda_{\max}-\lambda_{\min}\big)\,\eta_i,
\label{eq:mixing}
\end{equation}
where $\sigma(\cdot)$ is the logistic sigmoid, $\eta_i\in(0,1)$ is $z_i$ mapped to the unit interval, and the floor and ceiling of the mixing rate are $[\lambda_{\min},\lambda_{\max}]=[0.15,0.65]$ (implementation details and design rationale in Appendix~\ref{app:nta}). The choice of $\lambda_{\min}=0.15$ reflects a conservative strategy: even when the teacher provides little signal (low $\eta_i$), we retain some contrast weighting rather than discarding it entirely, to avoid complete reliance on response-level verifier advantages. This partial fallback allows fine-grained credit to be guided by a weak contrast while respecting the verifier's trajectory-level judgment. The ceiling $\lambda_{\max}=0.65$ prevents any single trajectory from being dominated by potentially noisy token contrasts. $\lambda_i$ varies per trajectory between this floor and ceiling; trajectories on which the teacher cannot discriminate the views receive low $\eta_i$ and reduced contrast weighting, while confidently resolved trajectories receive high $\eta_i$. Trajectories without a valid sibling pair (no negative hint) use single-view RLSD weights (Eq.~\eqref{eq:rlsd_gain}) at mixing rate $\lambda_{\min}$, a weaker fallback than the pairwise method; trajectories without any hint use uniform GRPO weights.

\subsection{Contrast-Weighted Credit Redistribution}
\label{sec:method_redistribution}

The token contrast enters the GRPO objective as a positive token weight built in three steps. First, the direction-aware log-weight incorporates the trajectory advantage sign to align the contrast with the reinforcement direction:
\begin{equation}
s_{i,t} = \operatorname{sign}(A_i)\, \frac{c_{i,t}}{\tau}
\label{eq:direction_aware_weight}
\end{equation}
with temperature $\tau$, so that tokens supporting the correct answer ($c_{i,t}>0$) are reinforced more on positive trajectories and penalized less on negative ones. The log-weight is then exponentiated and clipped:
\begin{equation}
\hat{w}_{i,t} = \operatorname{clip}\!\big( e^{s_{i,t}},\, u_{\min},\, u_{\max} \big),
\label{eq:raw_weight}
\end{equation}
with $\tau{=}1$ and lower/upper bounds $[u_{\min},u_{\max}]=[0.8,1.2]$, bounding the token-level weight ratio. Second, $\hat{w}_{i,t}$ is normalized to unit mean within each response, $\tilde{w}_{i,t} = \hat{w}_{i,t} \,\big/\, \big(\frac{1}{T_i}\sum_{t=1}^{T_i}\hat{w}_{i,t}\big)$, so credit is redistributed rather than amplified. Third, the normalized weight is mixed with the uniform weight $1$ at the mixing rate $\lambda_i$:
\begin{equation}
w^{\mathrm{mix}}_{i,t} = (1-\lambda_i) + \lambda_i\,\tilde{w}_{i,t},
\qquad
\hat{A}_{i,t} = A_i \cdot w^{\mathrm{mix}}_{i,t}.
\label{eq:final_advantage}
\end{equation}
Since $w^{\mathrm{mix}}_{i,t}>0$ and $\tilde{w}_{i,t}$ has unit mean, $w^{\mathrm{mix}}_{i,t}$ satisfies the constraints in Sec.~\ref{sec:preliminaries}, and the redistributed advantages replace $A_i$ in the GRPO surrogate:
\begin{equation}
\mathcal{J}_{\text{\scriptsize\method{}}}(\theta) = \mathbb{E}\!\left[ \tfrac{1}{G}\sum_{i,t} \tfrac{1}{T_i} \min\!\big( \rho_{i,t} \hat{A}_{i,t},\, \bar{\rho}_{i,t} \hat{A}_{i,t} \big) \right].
\label{eq:objective}
\end{equation}

\paragraph{Coefficient-level guarantees.}
Since $w^{\mathrm{mix}}_{i,t}>0$ and $\frac{1}{T_i}\sum_t w^{\mathrm{mix}}_{i,t}=1$, the token advantages satisfy two algebraic properties at the coefficient level: (1) mean preservation, $\frac{1}{T_i}\sum_t \hat{A}_{i,t} = A_i$, and (2) sign preservation, $\operatorname{sign}(\hat{A}_{i,t}) = \operatorname{sign}(A_i)$. These properties hold in the explicit formulas before policy optimization and ensure that the verifier reward determines whether the trajectory is reinforced or suppressed, while the token contrast redistributes credit among tokens within that fixed direction and magnitude. They do not extend to the clipped, non-linear PPO update in Eq.~\eqref{eq:objective}, whose per-token gradients also depend on the clipping regime (Appendix~\ref{app:sign_mean_preserving}). Algorithm~\ref{alg:training} in Appendix~\ref{app:implementation_details} summarizes one complete \method{} training step.

\begin{table*}[t]
\centering
\footnotesize
\setlength{\tabcolsep}{2.0pt}
\begin{tabularx}{\textwidth}{ll *{6}{Y}}
\toprule
Model & Method & GSM8K-V & LogicVista & MMMU & MathVision & MathVista & Average \\
\midrule
\multirow{8}{*}{\makecell{Qwen3-VL-8B\\Instruct}}
& Base  & 22.97 & 54.36 & 61.33 & 45.79 & 76.00 & 52.09 \\
& GRPO  & 26.61 & 58.84 & 65.89 & 47.11 & 75.70 & 54.83 ($\uparrow$2.74) \\
& PAPO  & 25.47 & 59.73 & 65.44 & 50.03 & 76.10 & 55.36 ($\uparrow$3.27) \\
& VPPO  & \underline{26.99} & \underline{61.52} & \underline{67.67} & \underline{52.80} & \underline{77.30} & \underline{57.25} ($\uparrow$5.16) \\
& OPSD  & 23.05 & 54.36 & 61.67 & 42.86 & 75.20 & 51.43 ($\downarrow$0.66) \\
& RLSD  & 26.38 & 57.72 & 66.22 & 49.90 & 76.70 & 55.38 ($\uparrow$3.29) \\
& RLCSD & 25.47 & 57.05 & 64.00 & 48.95 & 77.10 & 54.51 ($\uparrow$2.42) \\
& \cellcolor{blue!10}\textbf{\method{} (Ours)} 
& \cellcolor{blue!10}\textbf{28.13} 
& \cellcolor{blue!10}\textbf{63.31} 
& \cellcolor{blue!10}\textbf{69.00} 
& \cellcolor{blue!10}\textbf{54.01} 
& \cellcolor{blue!10}\textbf{77.90} 
& \cellcolor{blue!10}\textbf{58.47} ($\uparrow$6.38) \\
\midrule
\multirow{8}{*}{\makecell{Qwen2.5-VL-7B\\Instruct}}
& Base  & 15.96 & 45.08 & 51.28 & 25.52 & 67.80 & 41.13 \\
& GRPO  & 17.49 & 48.15 & 52.97 & 27.72 & 72.12 & 43.69 ($\uparrow$2.56) \\
& PAPO  & \textbf{18.33} & 46.36 & 53.83 & 27.42 & 70.97 & 43.38 ($\uparrow$2.26) \\
& VPPO  & 16.40 & \textbf{50.00} & \underline{55.53} & 27.61 & 71.30 & \underline{44.17} ($\uparrow$3.04) \\
& OPSD  & 16.38 & 47.15 & 53.25 & 26.49 & 69.12 & 42.48 ($\uparrow$1.35) \\
& RLSD  & 17.55 & 46.81 & 53.80 & \underline{28.90} & \underline{72.43} & 43.90 ($\uparrow$2.77) \\
& RLCSD & 16.24 & 45.02 & 52.19 & 26.13 & 68.42 & 41.60 ($\uparrow$0.48) \\
& \cellcolor{blue!10}\textbf{\method{} (Ours)}
& \cellcolor{blue!10}\underline{18.27}
& \cellcolor{blue!10}\underline{49.89}
& \cellcolor{blue!10}\textbf{55.67}
& \cellcolor{blue!10}\textbf{29.01}
& \cellcolor{blue!10}\textbf{73.00}
& \cellcolor{blue!10}\textbf{45.17} ($\uparrow$4.04) \\
\midrule
\multirow{8}{*}{InternVL3.5-4B}
& Base  & 14.76 & 49.13 & 61.76 & 33.60 & 70.47 & 45.95 \\
& GRPO  & 14.85 & 51.09 & 62.01 & \underline{34.89} & 71.58 & 46.88 ($\uparrow$0.93) \\
& PAPO  & 14.74 & 49.69 & 62.28 & 34.67 & 71.24 & 46.53 ($\uparrow$0.58) \\
& VPPO  & \underline{15.68} & \underline{52.52} & \underline{62.36} & 34.46 & \underline{72.03} & \underline{47.41} ($\uparrow$1.46) \\
& OPSD  & 14.53 & 52.27 & 61.03 & 33.78 & 71.03 & 46.53 ($\uparrow$0.58) \\
& RLSD  & \textbf{15.92} & 50.56 & 61.56 & 34.66 & 71.71 & 46.88 ($\uparrow$0.93) \\
& RLCSD & 14.69 & 49.22 & 61.55 & 33.75 & 70.37 & 45.92 ($\downarrow$0.03) \\
& \cellcolor{blue!10}\textbf{\method{} (Ours)}
& \cellcolor{blue!10}15.54
& \cellcolor{blue!10}\textbf{52.80}
& \cellcolor{blue!10}\textbf{63.78}
& \cellcolor{blue!10}\textbf{35.26}
& \cellcolor{blue!10}\textbf{72.60}
& \cellcolor{blue!10}\textbf{48.00} ($\uparrow$2.05) \\
\bottomrule
\end{tabularx}
\caption{Accuracy (\%) (mean@8) on five multimodal reasoning benchmarks. Parenthesized values denote percentage-point change from Base. Bold indicates best results; underline indicates second best.}
\label{tab:main_results}
\end{table*}

\section{Experiments}
\label{sec:experiments}

\subsection{Experimental Setup}
\label{sec:experimental_setup}

\paragraph{Baselines.} We compare \method{} with main multimodal RLVR methods. GRPO~\citep{shao2024deepseekmath} is the baseline that broadcasts a uniform sequence-level advantage to every token. OPSD~\citep{zhao2026self} trains the student policy by the privileged teacher with self-generated rollouts. RLSD~\citep{yang2026self} and RLCSD~\citep{pan2026rlcsd} are self-distilled RLVR methods that derive token-level credit from answer-conditioned teacher--student contrasts, and are the closest to our approach. PAPO~\citep{wang2026perception} and VPPO~\citep{huang2026spotlight} inject visually grounded signals into the RL objective, addressing grounding from a direction orthogonal to token credit.

\paragraph{Training.} \method{}, GRPO, RLSD, RLCSD, PAPO, and VPPO are trained under a shared configuration on the MMFineReason-123K training split by RLSD, which derives from MMFineReason~\citep{lin2026mmfinereason} and keeps only the final conclusion after the \texttt{</think>} tag as supervision. All methods are trained for 200 optimization steps with 8 rollouts per prompt (Appendix~\ref{tab:hyperparameters}). To evaluate scalability and generalizability across diverse architectures and model scales, we use Qwen3-VL-8B-Instruct~\citep{bai2025qwen3}, Qwen2.5-VL-7B-Instruct~\citep{bai2025qwen25vltechnicalreport}, and InternVL3.5-4B~\citep{wang2025internvl3} for training. OPSD uses its original implementation with rank-64 LoRA~\citep{hu2021lora} and the ms-swift trainer (Appendix~\ref{tab:opsd_config}). Checkpoint selection is based on the training reward over the final 50 steps.

\paragraph{Benchmarks.} We evaluate on five multimodal reasoning benchmarks. GSM8K-V~\citep{yuan2025gsm8k} renders grade-school math word problems as images, directly testing whether a model reads quantitative content from pixels. LogicVista~\citep{xiao2024logicvista} tests logical reasoning over visual scenes. MMMU~\citep{yue2024mmmu} is a college-level, multi-discipline multimodal understanding benchmark. MathVision~\citep{wang2024measuring} collects competition-level mathematical problems in visual contexts. MathVista~\citep{lu2024mathvista} aggregates a broad range of mathematical and visual reasoning tasks. All benchmarks are evaluated with VLMEvalKit~\citep{duan2024vlmevalkit}.

\subsection{Main Results}

\textbf{\method{} is best on every benchmark with 8B model.}
As shown in Table~\ref{tab:main_results}, on Qwen3-VL-8B-Instruct, \method{} ranks first on all five benchmarks and reaches 58.47\% on average, +6.38 over Base, +3.64 over GRPO, and +1.22 over the strongest baseline VPPO. The gains over Base are largest on LogicVista (+8.95), MathVision (+8.22), and MMMU (+7.67), and smallest on MathVista (+1.90), where all methods lie within 2.7 points of each other. VPPO is second on every benchmark, so the margin over it (0.60--1.79 points per benchmark) is the tightest comparison. \method{} reaches it at about a third of VPPO's per-step training time, as detailed in Appendix~\ref{app:training_time}. Appendix~\ref{app:case_study} visualizes \method{}'s token credit on three examples.

\textbf{Pairwise token contrast improves over teacher--student gaps.}
RLSD and RLCSD share \method{}'s training stack and differ mainly in how token credit is computed, so this comparison isolates the credit signal. \method{} outperforms RLSD by 3.09 points and RLCSD by 3.96 points on average, with the largest gaps on LogicVista (+5.59 over RLSD, +6.26 over RLCSD) and MathVision (+4.11 and +5.06). RLCSD, which contrasts correct and incorrect hints without the vision axis or adaptive trajectory confidence, gains 2.42 points over Base, fewer than GRPO (2.74), while \method{} gains 6.38. OPSD, trained with LoRA adapters in a separate framework, falls below Base, which we attribute to its different optimization setup rather than to on-policy distillation itself. The same pattern holds on the other two backbones: RLSD improves over GRPO by at most 0.55 points 
, while \method{} improves over GRPO by 3.64, 1.48, and 1.12 points.

\textbf{\method{} leads across different backbones.}
The advantage is not specific to Qwen3-VL-8B. On Qwen2.5-VL-7B-Instruct, \method{} achieves the best average accuracy (45.17\%, +4.04 over Base), ahead of VPPO (44.17\%), RLSD (43.90\%), and GRPO (43.69\%). It ranks first on MMMU, MathVision, and MathVista and is within 0.11 points of the best method on GSM8K-V and LogicVista. On InternVL3.5-4B, which comes from a different model family, \method{} again achieves the best average accuracy (48.00\%, +2.05 over Base) and ranks first on four of five benchmarks, with RLSD only marginally better on GSM8K-V (15.92 vs.\ 15.54). On both backbones, \method{} improves over Base on every benchmark rather than on a single task.

\textbf{The gain over baselines narrows with smaller self-teachers.}
The gain of \method{} over base model shrinks with backbone size: +6.38 points on Qwen3-VL-8B-Instruct, +4.04 points on Qwen2.5-VL-7B-Instruct, and +2.05 points on InternVL3.5-4B, while still remaining ahead of the strongest baseline. 
Since the self-teacher is a copy of the policy, a smaller backbone is less able to solve the hard training problems even with privileged hints, which makes its token contrasts noisier and limits how much token-level credit assignment can add over uniform credit.

\begin{figure*}[t]
\centering
\includegraphics[width=\textwidth]{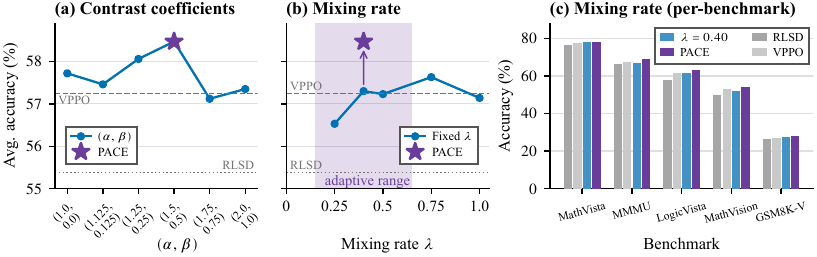}
\caption{Ablations on Qwen3-VL-8B-Instruct. (a) Contrast coefficients $(\alpha,\beta)$ with the adaptive mixing rate $\lambda_i$. (b) Fixed mixing rate $\lambda$. (c) Per-benchmark accuracy of fixed $\lambda$ and \method{}.}
\label{fig:ablation_sweep}
\end{figure*}

\subsection{Ablation Study}
\label{sec:ablation}

We ablate \method{} on Qwen3-VL-8B-Instruct and report average accuracy over the five benchmarks in Table~\ref{tab:ablation} and Figure~\ref{fig:ablation_sweep}; per-benchmark results are shown in Appendix~\ref{app:ablation_full}.

\paragraph{Pairwise token contrast is necessary.}
Table~\ref{tab:ablation}a compares hint contrasts, i.e.\ $\Delta_{i,t}$ or $\bar{\Delta}_{i,t}$ in Eq.~\eqref{eq:contrasts}, combined as full vision only or full $\pm$ null vision, against the \emph{vision gap} $\ell^{+,V}-\ell^{+,\varnothing}$, which uses no hint contrast; Fixed $\lambda$ sets $\lambda=0.5$ for all trajectories. Replacing hint contrasts with the vision gap drops accuracy to 51.05\%, below Base (52.09\%), with the largest loss on MathVision (40.95\% vs.\ 45.79\% for Base). 
With a fixed mixing rate, all hint-contrast variants reach 57.06--57.68\%, well above GRPO (54.83\%) and on par with VPPO (57.25\%). This 0.62 point spread is small beside the 6 point drop of the vision gap, so contrasting paired hints matters more than how the full- and null-vision contrasts are combined. Consistent with this, the vision gap drifts during training while the hint contrasts stay stable, as detailed in Appendix~\ref{app:training_signals}.

\begin{table}[t]
\centering
\small
\setlength{\tabcolsep}{2pt}
\begin{tabularx}{\columnwidth}{Xcc}
\toprule
\textbf{(a) Pairwise token contrast} & \textbf{Fixed $\lambda$} & \textbf{Adaptive $\lambda_i$} \\
\midrule
Vision gap (no hint contrast)       & 51.05 & -- \\
Hint contrast, full vision          & 57.06 & 57.72 \\
Hint contrast, full $-$ null vision & 57.68 & 57.34 \\
Hint contrast, full $+$ null vision & 57.23 & \cellcolor{blue!10}\textbf{58.47} \\
\midrule
\textbf{(b) Adaptive traj.\ confidence} & $\lambda$ & \textbf{Avg.}\\
\midrule
$S_{\mathrm{mean}}$ (mean-based)                      & [0.10,\,0.65] & 57.04 \\
$S_{\mathrm{mean}}$ (mean-based)                      & [0.15,\,0.75] & 57.42 \\
\rowcolor{blue!10}
$S_{\mathrm{mean}}$ (mean-based)                      & [0.15,\,0.65] & \textbf{58.47} \\
$S_{\mathrm{mean}}$ (mean-based)                      & [0.25,\,0.65] & 57.60 \\
$S_{\mathrm{std}}$ (dispersion-based)                & [0.25,\,0.65] & 57.62 \\
Mix of $S_{\mathrm{mean}}$ and $S_{\mathrm{std}}$      & [0.25,\,0.65] & 57.64 \\

\bottomrule
\end{tabularx}
\caption{Ablation results (average accuracy, \%). }
\label{tab:ablation}
\end{table}

\paragraph{Moderate counterfactual correction works best.}
Figure~\ref{fig:ablation_sweep}a varies the weights $(\alpha,\beta)$ of the two contrasts in Eq.~\eqref{eq:token_contrast} under adaptive mixing rate. Across $(\alpha,\beta)$, average accuracy 
peaks at the default $(1.5,0.5)$, followed by $(1.25,0.25)$. Removing the null-vision contrast $(1,0)$ or weighting it heavily ($\beta\ge0.75$) degrades accuracy.

\paragraph{Adaptive trajectory confidence adds gain beyond a fixed mixing rate.}
Figure~\ref{fig:ablation_sweep}b and c replace the adaptive $\lambda_i$ with a fixed rate shared by all trajectories. With full $+$ null vision contrast, adaptive trajectory confidence improves over fixed $\lambda=0.5$ by 1.24 points. Fixing $\lambda=0.40$, the median of the adaptive $\lambda_i$, reaches only 57.30\%, so 1.17 of the 1.24 points come from per-trajectory adaptation rather than from the average rate. No fixed rate exceeds 57.63\% (Figure~\ref{fig:ablation_sweep}b), and the best fixed rate varies across benchmarks (Figure~\ref{fig:ablation_sweep}c). The adaptive rate outperforms every fixed rate on GSM8K-V, MMMU, and MathVision.

\paragraph{The range of the mixing rate matters more than how confidence is measured.}
Table~\ref{tab:ablation}b varies the range $[\lambda_{\min},\lambda_{\max}]$ and the confidence $S_i$. Besides the default $S_{\mathrm{mean}}$ (Eq.~\eqref{eq:confidence}), the dispersion-based $S_{\mathrm{std}}=\tfrac12(\sigma^{+,\varnothing}_i+\sigma^{-,\varnothing}_i)-\tfrac12(\sigma^{+,V}_i+\sigma^{-,V}_i)$ measures how much more dispersed the teacher's token log-probabilities become without the image, where $\sigma^{a,b}_i$ is the within-trajectory standard deviation of $\ell^{a,b}_{i,t}$. Mix normalizes $S_{\mathrm{mean}}$ and $S_{\mathrm{std}}$ separately with Eq.~\eqref{eq:normalize}, giving $z^{\mathrm{mean}}_i$ and $z^{\mathrm{std}}_i$, and uses $\eta_i=\sigma\big(\tfrac12 z^{\mathrm{mean}}_i+\tfrac12 z^{\mathrm{std}}_i\big)$ in Eq.~\eqref{eq:mixing}. Relative to the default $[0.15,0.65]$, lowering the floor to $0.10$, raising it to $0.25$, or raising the ceiling to $0.75$ all lower accuracy: low-confidence trajectories still benefit from a moderate amount of contrast weighting, while high-confidence ones should not rely on it too heavily. At $[0.25,0.65]$, the three ways of measuring confidence perform similarly (57.60--57.64\%), so within the tested range the choice of $S_i$ matters less than the range of $\lambda_i$.

\section{Conclusion}

This paper presents \method{}, a method for self-distilled multimodal reasoning RLVR that assigns credit at two levels: pairwise token contrast decides which tokens of a response receive more credit, and adaptive trajectory confidence decides how far that credit is redistributed on each response.
Our experiments show that hint contrasts give more stable token credit than single-view signals, which drift during training. Normalizing trajectory confidence within each batch keeps the mixing rate stable throughout training.
With these designs, \method{} achieves the best average accuracy on evaluated benchmarks with models across different model families and model sizes from 4B to 8B. 

\newpage

\section*{Limitations}

Pairwise token contrast cancels only offsets shared by paired hints, and the null-vision intervention cannot rule out language shortcuts that correlate with the correct answer. The trajectory confidence that sets the mixing rate measures how well the teacher separates the full-information view from the degraded ones, not whether a response actually needs the image, so a high mixing rate does not guarantee visual grounding (Appendix~\ref{app:nta_limitations}). The sign- and mean-preserving properties hold for the token weights, not for the clipped PPO update. Gains are smaller on smaller backbones such as InternVL3.5-4B. On Qwen3-VL-2B-Instruct, \method{} does not achieve the best average (Appendix~\ref{app:qwen3vl2b}). We attribute this to the training data. Following RLSD, MMFineReason-123K keeps only problems that Qwen3-VL-4B-Thinking fails in all four rollouts, and on such hard problems smaller self-teachers remain unreliable even with privileged hints, so their token contrasts are noisier. Our experiments cover multimodal mathematical and logical reasoning with verifiable answers. Whether the gains transfer to other tasks remains to be tested. Finally, scoring four teacher views adds training cost, although inference is unaffected. 

\bibliography{custom}

\newpage
\appendix
\section{Related Works}
\label{app:extended_related_work}

\paragraph{Outcome-verifiable reinforcement learning.}
RLVR uses automatically checkable outcomes to avoid dense process supervision. GRPO estimates a relative advantage from multiple responses to one prompt without a separate value model~\citep{shao2024deepseekmath}, while subsequent systems study exploration, clipping, and training stability~\citep{guo2025deepseek}. Multimodal RLVR transfers this recipe to visual mathematics and general visual reasoning~\citep{huang2026vision,meng2025mm}. The common limitation relevant to \method{} is that a scalar response-level advantage is broadcast to every valid response token. \method{} keeps this verifier-defined direction and changes only the allocation of credit within the sampled trajectory.

\paragraph{On-policy distillation and self-distillation.}
On-policy distillation trains on prefixes and responses sampled by the student, reducing the mismatch between training trajectories and inference behavior~\citep{agarwal2024policy,zhao2026self}. Self-distilled RLVR methods instead let a single model play both roles: an EMA or periodically synchronized copy of the actor is given privileged context and used to score the same realized tokens. RLSD uses a privileged answer-conditioned teacher--student likelihood gap as a bounded token-credit signal while anchoring the update sign to the verifier~\citep{yang2026self}; SDPO instead conditions the self-teacher on rich textual feedback (e.g., an error explanation) rather than the ground-truth answer and distills the feedback-informed distribution densely into the student~\citep{hubotter2026reinforcement}; SRPO observes that self-distillation supervision is uninformative on already-correct rollouts and routes correct samples to GRPO while only incorrect samples receive dense self-distillation, with an entropy-aware weight on the teacher target~\citep{li2026unifying}. RLCSD contrasts correct and incorrect answer hints to suppress components shared by both conditions~\citep{pan2026rlcsd}. \method{} extends this line by asking whether the correctness contrast is specifically supported by the visual input.

\paragraph{Recent token-level credit assignment.}
Several contemporaneous papers also move beyond uniform sequence advantages. VCSD removes image content to create a visual contrast between original and content-erased inputs, then distills a sharpened full token distribution~\citep{liang2026visual}. GCPO contrasts predictions under positive and negative prompts and uses a predictive-divergence statistic to form per-token advantages~\citep{li2026guidance}. PEPO derives a perception prior from hidden-state similarity and combines it with token entropy through a smooth gate~\citep{li2026rethinking}. CEPO is the closest contemporaneous method: it also asks whether a token is favored by a correct answer hint \emph{and} disfavored by an incorrect one, built from an existing rejected rollout in the same batch, and uses the resulting agreement as a decisive-token signal on multimodal math reasoning~\citep{heakl2026cepo}. Orthogonally, a polarity--entropy analysis argues that token entropy upper-bounds how much hindsight credit a token can absorb and reallocates GRPO advantages by crossing this capacity term with reward polarity~\citep{he2026rethinking}. \method{} differs from all of these in three ways. First, its raw statistic is a four-view difference-in-differences: two correctness contrasts, $\Delta_{i,t}$ under observed vision and $\bar{\Delta}_{i,t}$ under nulled vision (Eq.~\eqref{eq:contrasts}), are combined into the token contrast $c_{i,t}$ (Eq.~\eqref{eq:token_contrast}), rather than relying on a single visual contrast, prompt divergence, perception--entropy score, or an answer contrast alone. Unlike CEPO, whose two teacher views vary only the answer hint with the image held fixed, \method{}'s $2\times2$ design crosses the answer hint with a full/null vision axis, so it can attribute the correctness contrast to the visual input specifically rather than to the hint alone. Second, the positive/negative views are answer candidates, so the statistic tests visual support for correctness rather than generic input sensitivity or token entropy. Third, $c_{i,t}$ is used as a detached signal for sign-preserving, mean-conserving GRPO token-level credit assignment; it is not itself a distillation target and is interpreted as a contrastive signal, not a causal effect.

\paragraph{Visual grounding and visual retention.}
PAPO and VPPO prioritize tokens with stronger visual dependence, and propagation-aware objectives study whether visual information survives along long multimodal chains~\citep{wang2026perception,huang2026spotlight,gong2026reflection}. These methods motivate measuring vision-aware token signals, but visual dependence is unsigned with respect to the answer: a wrong interpretation can also be highly image-sensitive. \method{} subtracts the visual sensitivity associated with an incorrect answer hint from the sensitivity associated with a correct hint, targeting correctness-aligned visual evidence.
\paragraph{Reflection and re-attention.}
Another family of multimodal RL methods adds reflection, search, or explicit re-attention to recover missing visual evidence~\citep{wang2026vl,wan2026srpo,yao2026mulberry,jian2025look,chu2025qwen}. Those approaches modify the reasoning process or introduce additional stages. \method{} is complementary: it scores the trajectory already sampled by the policy under controlled answer and visual contexts, and uses the resulting token contrast to assign credit without adding an inference-time reflection module.

\section{Implementation Details}
\label{app:implementation_details}

\subsection{Training Procedure}
\begin{algorithm}[t]
\caption{\method{} training step}
\label{alg:training}
\begin{algorithmic}[1]
\Require policy $\pi_\theta$, teacher $\bar\pi$ (copied from $\pi_\theta$ every 20 steps), rule-based verifier
\State Sample $G$ responses $\{y_i\}_{i=1}^{G}$ per prompt from $\pi_{\theta_{\mathrm{old}}}$; compute rewards $r_i$ and advantages $A_i$ 
\State Build hints: $h^{+}_i$ from ground truth; $\{h^{-}_{i,k}\}$ from incorrect siblings
\State Teacher-force each $y_i$ under the four teacher views \Comment{Eq.~\eqref{eq:views}}
\State Pairwise token contrast: compute token contrast $c_{i,t}$ \Comment{Eqs.~\eqref{eq:contrasts}--\eqref{eq:token_contrast}}
\State Adaptive trajectory confidence: confidence $S_i$, $z_i$, $\eta_i$, $\lambda_i$ \Comment{Eqs.~\eqref{eq:confidence}--\eqref{eq:mixing}}
\State Redistribute: $\hat{A}_{i,t} = A_i \cdot w^{\mathrm{mix}}_{i,t}$ \Comment{Eqs.~\eqref{eq:raw_weight}--\eqref{eq:final_advantage}}
\State PPO-clipped policy update on $\mathcal{J}_{\method{}}$ \Comment{Eq.~\eqref{eq:objective}}
\end{algorithmic}
\end{algorithm}

Algorithm~\ref{alg:training} walks through one \method{} optimization step end to end. Given a batch of prompts, the old policy $\pi_{\theta_{\mathrm{old}}}$ samples $G$ responses per prompt, and the rule-based verifier scores them to form group-relative advantages $A_i$ (Sec.~\ref{sec:preliminaries}). We then build the answer hints: a single positive hint $h^{+}_i$ from the ground-truth answer, and $K_i$ negative hints $\{h^{-}_{i,k}\}$ collected from incorrect sibling rollouts in the same group and deduplicated by their normalized final answers (Sec.~\ref{sec:method_views}). Each sampled response is teacher-forced under the four teacher views --- correct vs.\ incorrect hints crossed with full vs.\ null vision --- producing four detached per-token log-probability sequences from the teacher. Pairwise token contrast over these views defines the token contrast $c_{i,t}$ (Sec.~\ref{sec:method_contrast}), while adaptive trajectory confidence, i.e.\ the teacher confidence $S_i$, its batch-normalized value $z_i$, the sigmoid output $\eta_i$, and the mixing rate $\lambda_i$, decides how much the token contrast is trusted on each trajectory (Sec.~\ref{sec:method_nta}). The group advantage is then redistributed across tokens as $\hat{A}_{i,t} = A_i \cdot w^{\mathrm{mix}}_{i,t}$ (Sec.~\ref{sec:method_redistribution}), and the policy is updated on the PPO-clipped objective of Eq.~\eqref{eq:objective}. Trajectories without a valid negative-hint sibling fall back to single-view RLSD weights at mixing rate $\lambda_{\min}$, and those without any hint use uniform GRPO weights. All teacher scoring is detached, computed once per step under teacher forcing, and discarded after the update, so the procedure adds no inference-time cost at deployment.

\subsection{Training Configurations}
\method{} and the other methods trained under the shared setup are implemented on top of the open-source RLSD codebase, which itself builds on the veRL reinforcement learning framework~\citep{sheng2025hybridflow}; Table~\ref{tab:hyperparameters} summarizes this shared training configuration. For all methods in the shared stack, the student is the respective backbone (Qwen3-VL-8B-Instruct, Qwen3-VL-2B-Instruct~\citep{bai2025qwen3}, Qwen2.5-VL-7B-Instruct~\citep{bai2025qwen25vltechnicalreport}, or InternVL3.5-4B~\citep{wang2025internvl3}) fine-tuned in full parameters, and the teacher is a BF16 copy of the policy that starts from the base model and is hard-refreshed from the actor every 20 steps for 8B, 7B, and 4B models, following RLSD.

\paragraph{Qwen3-VL-2B-Instruct configuration.}
The 2B model uses a tighter teacher synchronization interval and a narrower range of the mixing rate to account for faster policy drift in smaller models. Specifically, the teacher is hard-refreshed every 10 steps (instead of 20), because the 2B policy shifts more rapidly during training and a 20-step-old teacher becomes too stale to provide reliable token contrasts. The mixing rate $\lambda_i$ is adapted within $[0.05, 0.30]$ (instead of $[0.15, 0.65]$), reducing reliance on potentially noisier token contrasts while retaining a modest degree of token-level credit assignment. All other hyperparameters remain identical to Table~\ref{tab:hyperparameters}.

The OPSD baseline is trained separately with ms-swift~\citep{zhao2024swiftascalablelightweightinfrastructure} and its generalized knowledge distillation (GKD) trainer (Table~\ref{tab:opsd_config}). Following the original OPSD configuration~\citep{zhao2026self}, OPSD uses LoRA adapters of rank 64 and $\alpha=128$ on all linear layers, with the teacher as the same frozen base model accessed by disabling the adapters. OPSD employs forward KL between teacher and student distributions (GKD $\beta=0$), fully on-policy student rollouts (GKD $\lambda=1.0$), sampling temperature 1.1, no SFT regularization, learning rate $5\times10^{-6}$, and batch size 32. While we maintain 200 optimization steps and the same MMFineReason-123K training split for comparability, the different framework and hyperparameters reflect OPSD's original implementation.

\begin{table}[t]
\centering
\small
\renewcommand{\arraystretch}{1.08}
\begin{tabularx}{\columnwidth}{@{} l >{\raggedright\arraybackslash}X @{}}
\toprule
\textbf{Configuration} & \textbf{Value} \\
\midrule
Training data & MMFineReason-123K \\
Optimization steps & 200 \\
Batch size & 256 prompts \\
Rollouts per prompt & 8 \\
Actor global batch size & 256 \\
Rollout temperature & 1.0 \\
Max training response length & 4,096 \\
Optimizer & AdamW \\
Learning rate & $1\times10^{-6}$ \\
Weight decay & 0.01 \\
Policy-ratio bounds & $[0.80, 1.28]$ \\
Reference KL & Disabled \\
Teacher type & Policy copy, hard-refreshed every 20 steps, BF16 \\
Negative hints $K$ & 1 \\
Contrast coefficients $(\alpha,\beta)$ & $(1.5,\,0.5)$ \\
Contrast temperature $\tau$ & 1.0 \\
Raw log-weight clip $s_{\max}$ & 8.0 \\
Pre-norm weight range & $[0.8, 1.2]$ \\
Mixing rate $\lambda_i$ & adaptive in $[0.15, 0.65]$ \\
Hardware & 8 $\times$ NVIDIA H200 \\
\bottomrule
\end{tabularx}
\caption{Training configuration of \method{} and the methods trained under the shared setup.}
\label{tab:hyperparameters}
\end{table}

\begin{table}[t]
\centering
\small
\renewcommand{\arraystretch}{1.1}
\begin{tabularx}{\columnwidth}{@{} l >{\raggedright\arraybackslash}X @{}}
\toprule
\textbf{Configuration} & \textbf{Value} \\
\midrule
Training data & MMFineReason-123K \\
Optimization steps & 200 \\
Batch size & 32 \\
Optimizer & AdamW \\
Learning rate & $5\times10^{-6}$ \\
Warmup ratio & 0 \\
Precision & BF16 \\
LoRA rank / $\alpha$ & 64 / 128 \\
LoRA target modules & all-linear \\
Frozen modules & ViT and aligner \\
Divergence & Forward KL (GKD $\beta=0$) \\
On-policy fraction (GKD $\lambda$) & 1.0 \\
Sampling temperature & 1.1 \\
SFT loss weight & 0 \\
Max sequence length & 8,192 \\
Max completion length & 4,096 \\
Teacher & Frozen base model (adapters disabled) \\
Generation backend & vLLM (colocate mode) \\
DeepSpeed & ZeRO-2 \\
Hardware & 4 $\times$ NVIDIA H200 \\
\bottomrule
\end{tabularx}
\caption{OPSD baseline configuration with the ms-swift GKD trainer.}
\label{tab:opsd_config}
\end{table}

\subsection{Teacher Prompt}
\label{app:teacher_prompt}

The positive and negative teacher branches share an identical prompt template and differ only in the candidate answer content. Below is the positive-branch prompt: the question (with the image placeholder) followed by the ground-truth answer as a privileged candidate, wrapped by the chat template. The negative branch reuses the identical template with an incorrect candidate answer parsed from a sibling rollout. The visual-null views replace the image embedding with zeros while keeping this prompt, including the image token positions, unchanged.

\begin{tcblisting}{
    colframe=black,
    colback=gray!15,
    listing only,
    breakable,
    listing options={
        breaklines=true,
        basicstyle=\ttfamily\small,
        columns=fullflexible
    }
}
<|im_start|>user
<image>
{{ question | trim }}

Privileged candidate answer: {{ hint | trim }}

Use the candidate answer as privileged context while independently solving the
original problem. You FIRST think about the reasoning process as an internal
monologue enclosed within <thought> </thought> tags. The final answer MUST be
enclosed within <answer> </answer> tags.<|im_end|>
<|im_start|>assistant
\end{tcblisting}

\subsection{Shared-Offset Cancellation in Pairwise Token Contrast}
\label{app:shared_offset}

This appendix states precisely which offsets the token contrast $c_{i,t}$ (Eq.~\eqref{eq:token_contrast}) removes, and which it does not.

\paragraph{Cancellation of shared offsets.}
Let $\ell = \log\bar\pi(y_{i,t}|x, h, y_{i,<t})$ denote a teacher log-probability under some view, with $h$ the hint of that view. As the student moves, the teacher $\bar\pi$, fixed between refreshes, scores increasingly off-distribution responses, which may shift its log-probabilities. Suppose, as an idealization, that this shift is a common offset for both hint views: $\ell' = \ell + \delta$.

Each correctness contrast $\Delta_{i,t}$ compares positive and negative hints under the same visual context:
\[
\Delta'_{i,t} = \big(\ell^{+,V}_{i,t} + \delta\big) - \big(\ell^{-,V}_{i,t} + \delta\big) = \Delta_{i,t}.
\]
The shift $\delta$ cancels because both branches of the contrast are affected equally. The same cancellation holds for $\bar{\Delta}_{i,t}$ under the nulled-vision branch. Therefore, the combined token contrast $c_{i,t} = \alpha\Delta_{i,t} + \beta\bar{\Delta}_{i,t}$ preserves the \emph{relative correctness signal} while removing any offset shared by the paired hint views.

\paragraph{Scope.}
This is an algebraic property of the contrast, not a claim about training dynamics. It removes only offsets common to both branches of a contrast. The teacher's uncertainty may shift non-uniformly: the positive and negative hint branches may respond differently to off-distribution responses, and nulled-vision views may shift differently from full-vision ones. Such view-specific shifts are not removed, and we do not claim that $c_{i,t}$ is robust to policy drift in general.

\paragraph{What the cancellation does and does not imply.}
The cancellation does \emph{not} mean that $\mathbb{E}_t[c_{i,t}]=0$ for a single trajectory or that the token contrast is zero-mean in a statistical sense. Instead, it means:
\begin{itemize}[leftmargin=*,topsep=2pt,itemsep=1pt]
\item The token contrast $c_{i,t}$ contains no offset that is shared by the paired hint views, because every view appears inside a contrast and such terms cancel in the subtraction.
\item The statistic is a relative quantity rather than a zero-mean one: it measures relative correctness support (the difference between how strongly the token supports the correct vs.\ incorrect answer), not an absolute confidence level.
\item More formally: if we decompose each log-probability into a trajectory-level mean and token deviation, $\ell^{+,V}_{i,t} = \mu^{+,V} + \delta^{+,V}_{i,t}$ with trajectory mean $\mu^{+,V}$ (the $\mu^{\cdot}_i$ of Eq.~\eqref{eq:confidence}, index $i$ dropped) and token deviation $\delta^{+,V}_{i,t}$, and analogously for the other views, then $c_{i,t}$ depends only on the \emph{differences} $(\mu^{+,V} - \mu^{-,V})$ and $(\mu^{+,\varnothing} - \mu^{-,\varnothing})$, not on the absolute levels $\mu^{+,V}$ or $\mu^{-,V}$ themselves.
\end{itemize}
The construction thus removes shared offsets and retains the correctness-aligned contrast; whether this reduces sensitivity to policy drift in practice depends on how much of the drift is shared across paired views, which we examine below at the batch level. The token contrast can still be positive or negative on average across a trajectory, depending on whether the response as a whole supports the correct answer more than the incorrect one.

\paragraph{Empirical verification.}
Figure~\ref{fig:training_signals}a tracks these quantities during training. The vision gap, a single-view signal, rises from 0.20 to 0.42 (first vs.\ last 20 steps), while the batch means of $\Delta$ and $\bar{\Delta}$ stay within 0.02 of zero and shift by at most 0.011; in a separate RLSD run (not plotted), the teacher--student gap drifts from $-0.019$ to $-0.055$. At the batch level, most of the drift is therefore shared by the paired hint views, as the cancellation argument requires. These are batch means, so they do not rule out token- or view-specific shifts; measuring the variance of $c_{i,t}$ across steps relative to that of the individual view log-probabilities remains future work.

\subsection{Adaptive Trajectory Confidence: Implementation Details}
\label{app:nta}

This appendix specifies how the normalized confidence $z_i$ in Eq.~\eqref{eq:normalize} is implemented and the rationale for making the normalization batch-relative and robust.

\paragraph{Estimator.}
Within each actor micro-batch $\mathcal{B}$, let $\mathcal{V}\subseteq\mathcal{B}$ be the rows with a valid hint pair. In the actual implementation, the denominator of Eq.~\eqref{eq:normalize} is rescaled by the constant $1.4826$ and stabilized by a small constant $\epsilon$:
\begin{equation*}
z_i = \operatorname{clip}\!\Big( \frac{S_i - \operatorname{med}_{\mathcal{V}}(S)}{1.4826\,\operatorname{MAD}_{\mathcal{V}}(S) + \epsilon},\, -\zeta,\, \zeta \Big).
\end{equation*}
The factor $1.4826\approx1/\Phi^{-1}(3/4)$, where $\Phi^{-1}$ is the standard normal quantile function, makes $1.4826\,\operatorname{MAD}$ a consistent estimator of the standard deviation when $S$ is normally distributed, so $z_i$ lives on the same scale as an ordinary $z$-score and the clip bound $\zeta{=}3$ retains its usual ``three-sigma'' meaning. The constant $\epsilon$ only prevents division by zero when nearly all values of $S$ in $\mathcal{V}$ coincide and the MAD vanishes. Three boundary cases are handled explicitly:
\begin{enumerate}[leftmargin=*,topsep=2pt,itemsep=1pt]
\item If a batch has no reward variance ($\operatorname{std}(\{r_j\}_{j\in\mathcal{B}})=0$), all responses are equivalent; we set $A_i=0$ for all $i$ and skip the RLVR update.
\item If a response has no valid negative sibling ($K_i=0$, so the sum in Eq.~\eqref{eq:views} is undefined), we use single-view RLSD weights (Eq.~\eqref{eq:rlsd_gain}) at mixing rate $\lambda_i=\lambda_{\min}$, a weaker fallback than the pairwise method.
\item If $|\mathcal{V}|<4$ (too few valid hint pairs for a stable normalization), every valid row receives the neutral value $\eta_i=0.5$; rows without any hint use uniform GRPO credit.
\end{enumerate}

\paragraph{Why batch-relative: tracking the current level.}
The absolute level of the confidence $S_i$ changes over training, because the policy --- and hence the distribution of sampled responses that the teacher scores --- keeps moving. An absolute criterion is therefore fragile: in preliminary experiments, a fixed threshold on $S$ collapsed, with the mixing rate saturating once the overall level shifted. The batch median acts as the \emph{current water level}: the normalization only answers whether a trajectory is strong or weak \emph{relative to its current batch}, so a shift common to the whole batch does not change $z_i$.

\paragraph{Why median/MAD: heavy tails.}
The distribution of $S_i$ is heavy-tailed --- on a few trajectories the teacher is extremely confident and produces outlier values of $S_i$. Standardizing with mean/std lets those outliers inflate the scale and compresses the $z$-scores of the vast majority of ordinary trajectories into a narrow band, destroying the resolution of the normalized confidence exactly where it matters. The median and MAD are robust to such outliers, so the bulk of the distribution keeps full discriminative resolution.

\paragraph{Why clipping and the sigmoid midpoint.}
Clipping $z_i$ to $[-3,3]$ keeps the sigmoid away from its saturating regimes, so $\lambda_i$ is never pinned to the interval endpoints; throughout training we monitor the quantiles of $\eta_i$, and the observed P10/P90 of $0.29/0.83$ confirm that the sigmoid mapping operates in its responsive regime. The midpoint carries an explicit semantics: a trajectory at the batch median ($z_i=0$) receives $\eta_i=0.5$, i.e.\ $\lambda_i=0.4$, the midpoint of $[0.15,0.65]$ --- a typical trajectory receives a typical mixing rate.

\paragraph{Why micro-batch rather than global statistics.}
The four teacher forward passes for a micro-batch are already in hand when the normalization is computed, so the standardization adds zero communication and negligible compute --- measured at $0.1$--$0.2$\,s per training step, about $0.05\%$ of the step time. A global running estimate would add synchronization across data-parallel ranks and, worse, would compare a trajectory against a stale level that lags the shifting distribution, partially reintroducing the sensitivity to drift that batch-relative standardization avoids.

\paragraph{Limitations: confidence vs.\ visual necessity.}
\label{app:nta_limitations}
$S_i$ reflects the teacher's full-information confidence rather than whether the response actually needs the image. A response on which the teacher is confident due to the correct hint alone will still receive a fairly high $S_i$: removing the image lowers $\mu^{+,\varnothing}_i$ only slightly, so only the $\mu^{-,V}_i$ and $\mu^{-,\varnothing}_i$ terms separate the full-information view from the rest. Two points qualify what adaptive trajectory confidence provides:
\begin{enumerate}[leftmargin=*,topsep=2pt,itemsep=1pt]
\item If the image is irrelevant to a response, then $\Delta_{i,t}\approx\bar{\Delta}_{i,t}$ and the token contrast reduces to $c_{i,t}\approx(\alpha+\beta)\Delta_{i,t}$, a pure answer-correctness contrast. A high mixing rate on such a trajectory still redistributes credit by correctness, but not by visual grounding.
\item Adaptive trajectory confidence therefore acts as a \emph{plausibility check} on the teacher's overall quality rather than a guarantee of visual grounding: it down-weights the token contrast mainly on trajectories where the teacher barely separates the full-information view from the degraded ones.
\end{enumerate}

\subsection{Sign- and Mean-Preserving Properties: Scope and Limitations}
\label{app:sign_mean_preserving}

The main text claims that the token-level advantages $\hat{A}_{i,t}$ satisfy two algebraic properties: (1) per-trajectory mean preservation, $\frac{1}{T_i}\sum_t \hat{A}_{i,t} = A_i$, and (2) sign preservation, $\operatorname{sign}(\hat{A}_{i,t}) = \operatorname{sign}(A_i)$. These properties hold \emph{at the coefficient level}, i.e., in the explicit formulas before policy optimization. Specifically:
\begin{itemize}[leftmargin=*,topsep=2pt,itemsep=1pt]
\item The normalization step $\tilde{w}_{i,t} = \hat{w}_{i,t} \,/\, \bar{w}_i$ (where $\bar{w}_i = \frac{1}{T_i}\sum_t \hat{w}_{i,t}$) ensures that $\frac{1}{T_i}\sum_t w^{\mathrm{mix}}_{i,t} = 1$.
\item Since $A_i$ is a scalar multiplier on all $\hat{A}_{i,t}$, the mean property follows: $\frac{1}{T_i}\sum_t \hat{A}_{i,t} = A_i \cdot 1 = A_i$.
\item The mixing formula $w^{\mathrm{mix}}_{i,t} = (1-\lambda_i) + \lambda_i \tilde{w}_{i,t}$ is always positive (since both $\lambda_i$ and $\tilde{w}_{i,t}$ are positive), so $\operatorname{sign}(\hat{A}_{i,t}) = \operatorname{sign}(A_i)$.
\end{itemize}

However, the GRPO objective (Eq.~\eqref{eq:objective}) contains $\min(\rho_{i,t} \hat{A}_{i,t}, \bar{\rho}_{i,t} \hat{A}_{i,t})$ and implicit policy gradient clipping. These are \emph{non-linear} operations. Therefore, after policy optimization:
\begin{itemize}[leftmargin=*,topsep=2pt,itemsep=1pt]
\item The actual gradient flow to individual tokens depends on which term in the $\min(\cdot)$ is active (a function of importance weights $\rho_{i,t}$ and $\bar{\rho}_{i,t}$).
\item The magnitude of token-level updates is not directly controlled by the coefficient structure; it depends on the clipping regime and the distribution of advantage magnitudes.
\item The properties guarantee that the \emph{direction and sign} of credit assignment remain aligned with the trajectory-level verifier reward, and that credit is \emph{redistributed} (not amplified) among tokens. But they do not guarantee that the final update magnitudes will preserve the exact mean or relative weighting structure after clipping.
\end{itemize}

In summary, the sign- and mean-preserving properties hold at the coefficient level and ensure that the verifier reward controls the overall optimization direction. The token contrast cannot flip the sign or amplify the trajectory advantage. However, the effect size of individual token updates within a trajectory is not fully determined by the coefficient structure and depends on the policy gradient dynamics.

\subsection{Training Dynamics}
\label{app:training_dynamics}

\begin{figure}[t]
    \centering
    \includegraphics[width=\columnwidth]{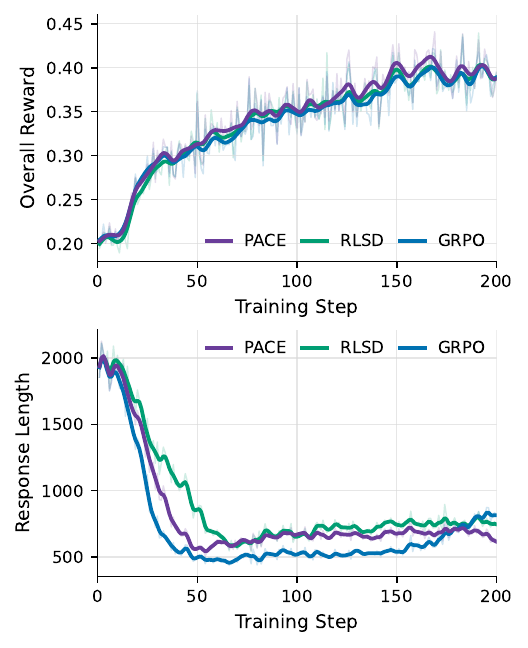}
    \caption{Training dynamics of \method{}, RLSD, and GRPO on Qwen3-VL-8B-Instruct over the shared 200-step budget. Top: overall training reward, i.e.\ the fraction of sampled responses passing the verifier. Bottom: mean response length in tokens. Light traces show raw values and bold curves show smoothed trends.}
    \label{fig:reward_length}
\end{figure}

Figure~\ref{fig:reward_length} tracks the overall training reward and the mean response length of \method{}, RLSD, and GRPO over the 200-step budget. The reward curves are close throughout: they rise from about 0.20 to 0.39, \method{} is slightly ahead between roughly steps 120 and 170 (peaking near 0.41), and the three converge again by step 200. The training reward therefore does not separate the methods; the differences in Table~\ref{tab:main_results} appear only on held-out benchmarks.

Response length orders the methods more clearly. All three compress responses sharply during the first 50 steps, from about 2{,}000 tokens to 500--900. GRPO compresses fastest, stays near 500 tokens until step 150, and then re-inflates to about 800 tokens by step 200. RLSD compresses slowest and stays above \method{} from about step 110 on. \method{} compresses at an intermediate rate, remains flat at 600--700 tokens, and ends with the shortest responses (about 620 tokens at step 200). The curves describe rather than explain the effect of token credit; the component-level evidence comes from the ablations (Table~\ref{tab:ablation}).

\subsection{Teacher Signals During Training}
\label{app:training_signals}

\begin{figure}[t]
\centering
\includegraphics[width=\columnwidth]{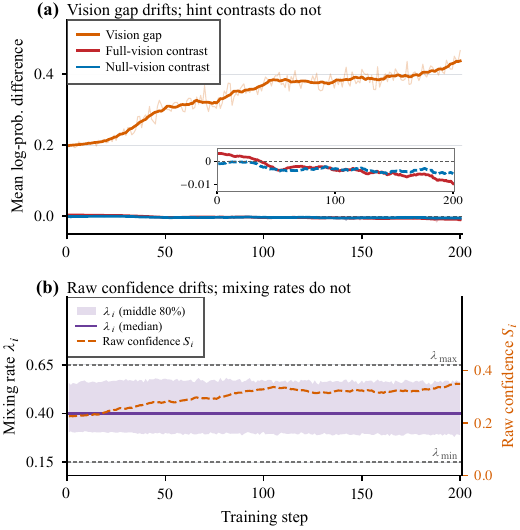}
\caption{Teacher signals of \method{} training on Qwen3-VL-8B-Instruct. (a)~Vision gap and hint contrasts (full-vision and null-vision). Inset: zoomed view of the hint contrasts. (b)~Mixing rate $\lambda_i$ (median and P10--P90 band) and raw confidence $S_i$.}
\label{fig:training_signals}
\end{figure}

\paragraph{Vision gap drifts but pairwise token contrast does not.}
Figure~\ref{fig:training_signals}a tracks the vision gap and hint contrasts during training. The vision gap doubles from 0.20 to 0.42 as the teacher improves: its full-vision log-probabilities pull away from null-vision ones, and this shift enters the signal directly regardless of whether a token depends on the image. The hint contrasts stay within 0.02 of zero and move by at most 0.011, about 20$\times$ less than the vision gap. The drift is shared by both hint branches of each contrast and cancels in subtraction. A 0.4 drift produces weight factors $e^{0.4}\approx1.49$ that exceed the clip bound $u_{\max}$, flattening token differences and explaining why the vision-gap variant falls below Base.

\paragraph{Adaptive trajectory confidence is batch-relative.}
Figure~\ref{fig:training_signals}b shows that while the raw confidence $S_i$ rises from 0.23 to 0.34 as the teacher grows more confident, the distribution of the mixing rate $\lambda_i$ remains stable: the P10--P90 band stays at 0.30--0.57 and the median at 0.40 throughout training. The robust batch-relative standardization absorbs both the level and scale drift of $S_i$, so adaptive trajectory confidence ranks trajectories by confidence relative to the current batch rather than shifting credit toward the token contrast as training progresses.

\subsection{Training-Time Cost Analysis}
\label{app:training_time}

\begin{figure}[h]
    \centering
    \includegraphics[width=\linewidth]{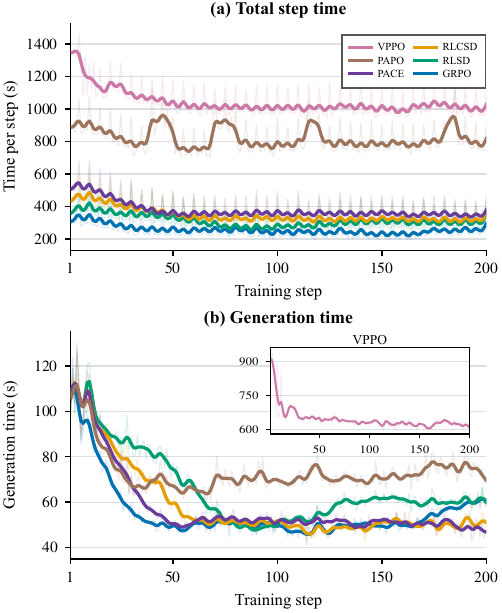}
    \caption{Per-step training and generation time across RLVR methods on an 8 $\times$ NVIDIA H200 GPU cluster. (a) Total step time. (b) Generation time.}
    \label{fig:training_time}
\end{figure}

Figure~\ref{fig:training_time} reports the empirical per-step training and generation time across RLVR methods on an 8 $\times$ NVIDIA H200 GPU cluster. The pairwise token contrast of \method{} requires positive and negative teacher contexts under both full and nulled vision. With one negative hint, this corresponds to four scored contexts per response, implemented as sequential microbatches to bound memory use. After the first 50 steps, a \method{} step takes about 350--380\,s, compared with about 250--280\,s for GRPO and 300--350\,s for RLSD and RLCSD, and far less than PAPO (about 800\,s) and VPPO (about 1{,}000\,s). Generation time (panel b) is close to that of GRPO (about 50\,s per step), so the overhead comes from teacher scoring rather than from longer rollouts. The additional computational work is confined strictly to training; evaluation and deployment use the original policy alone without any overhead.

\section{Experimental Details}

\subsection{Main Results with Standard Deviation}
\label{app:main_std}

Table~\ref{tab:main_std} reports per-benchmark accuracy (mean@8) with standard deviation on Qwen3-VL-8B-Instruct, Qwen2.5-VL-7B-Instruct, InternVL3.5-4B, and Qwen3-VL-2B-Instruct.

\begin{table*}[t]
\centering
\footnotesize
\setlength{\tabcolsep}{1.5pt}
\begin{tabularx}{\textwidth}{ll *{6}{Y}}
\toprule
Model & Method & GSM8K-V & LogicVista & MMMU & MathVision & MathVista & Average \\
\midrule
\multirow{8}{*}{\makecell{Qwen3-VL-8B\\Instruct}}
& Base  & 22.97{\scriptsize$\pm$0.63} & 54.36{\scriptsize$\pm$1.11} & 61.33{\scriptsize$\pm$0.68} & 45.79{\scriptsize$\pm$0.44} & 76.00{\scriptsize$\pm$0.72} & 52.09 \\
& GRPO  & 26.61{\scriptsize$\pm$0.70} & 58.84{\scriptsize$\pm$1.24} & 65.89{\scriptsize$\pm$1.18} & 47.11{\scriptsize$\pm$0.57} & 75.70{\scriptsize$\pm$0.96} & 54.83 ($\uparrow$2.74) \\
& PAPO  & 25.47{\scriptsize$\pm$0.64} & 59.73{\scriptsize$\pm$1.10} & 65.44{\scriptsize$\pm$0.73} & 50.03{\scriptsize$\pm$0.47} & 76.10{\scriptsize$\pm$0.74} & 55.36 ($\uparrow$3.27) \\
& VPPO  & \underline{26.99{\scriptsize$\pm$0.46}} & \underline{61.52{\scriptsize$\pm$0.94}} & \underline{67.67{\scriptsize$\pm$0.66}} & \underline{52.80{\scriptsize$\pm$0.40}} & \underline{77.30{\scriptsize$\pm$0.69}} & \underline{57.25} ($\uparrow$5.16) \\
& OPSD  & 23.05{\scriptsize$\pm$0.76} & 54.36{\scriptsize$\pm$1.04} & 61.67{\scriptsize$\pm$0.70} & 42.86{\scriptsize$\pm$0.60} & 75.20{\scriptsize$\pm$0.48} & 51.43 ($\downarrow$0.66) \\
& RLSD  & 26.38{\scriptsize$\pm$0.27} & 57.72{\scriptsize$\pm$1.26} & 66.22{\scriptsize$\pm$0.98} & 49.90{\scriptsize$\pm$0.49} & 76.70{\scriptsize$\pm$0.73} & 55.38 ($\uparrow$3.29) \\
& RLCSD & 25.47{\scriptsize$\pm$0.70} & 57.05{\scriptsize$\pm$1.62} & 64.00{\scriptsize$\pm$0.70} & 48.95{\scriptsize$\pm$0.82} & 77.10{\scriptsize$\pm$0.78} & 54.51 ($\uparrow$2.42) \\
& \cellcolor{blue!10}\textbf{\method{} (Ours)}
& \cellcolor{blue!10}\textbf{28.13{\scriptsize$\pm$0.52}}
& \cellcolor{blue!10}\textbf{63.31{\scriptsize$\pm$1.52}}
& \cellcolor{blue!10}\textbf{69.00{\scriptsize$\pm$0.87}}
& \cellcolor{blue!10}\textbf{54.01{\scriptsize$\pm$0.42}}
& \cellcolor{blue!10}\textbf{77.90{\scriptsize$\pm$0.83}}
& \cellcolor{blue!10}\textbf{58.47} ($\uparrow$6.38) \\
\midrule
\multirow{8}{*}{\makecell{Qwen2.5-VL-7B\\Instruct}}
& Base  & 15.96{\scriptsize$\pm$0.28} & 45.08{\scriptsize$\pm$1.58} & 51.28{\scriptsize$\pm$0.72} & 25.52{\scriptsize$\pm$0.43} & 67.80{\scriptsize$\pm$0.49} & 41.13 \\
& GRPO  & 17.49{\scriptsize$\pm$0.28} & 48.15{\scriptsize$\pm$0.64} & 52.97{\scriptsize$\pm$0.66} & 27.72{\scriptsize$\pm$0.36} & 72.12{\scriptsize$\pm$0.39} & 43.69 ($\uparrow$2.56) \\
& PAPO  & \textbf{18.33{\scriptsize$\pm$0.13}} & 46.36{\scriptsize$\pm$1.87} & 53.83{\scriptsize$\pm$1.02} & 27.42{\scriptsize$\pm$0.34} & 70.97{\scriptsize$\pm$0.60} & 43.38 ($\uparrow$2.26) \\
& VPPO  & 16.40{\scriptsize$\pm$0.52} & \textbf{50.00{\scriptsize$\pm$1.62}} & \underline{55.53{\scriptsize$\pm$0.50}} & 27.61{\scriptsize$\pm$0.19} & 71.30{\scriptsize$\pm$0.29} & \underline{44.17} ($\uparrow$3.04) \\
& OPSD  & 16.38{\scriptsize$\pm$0.77} & 47.15{\scriptsize$\pm$1.63} & 53.25{\scriptsize$\pm$0.90} & 26.49{\scriptsize$\pm$0.82} & 69.12{\scriptsize$\pm$0.26} & 42.48 ($\uparrow$1.35) \\
& RLSD  & 17.55{\scriptsize$\pm$0.26} & 46.81{\scriptsize$\pm$0.64} & 53.80{\scriptsize$\pm$0.52} & \underline{28.90{\scriptsize$\pm$0.32}} & \underline{72.43{\scriptsize$\pm$0.38}} & 43.90 ($\uparrow$2.77) \\
& RLCSD & 16.24{\scriptsize$\pm$0.28} & 45.02{\scriptsize$\pm$2.24} & 52.19{\scriptsize$\pm$1.02} & 26.13{\scriptsize$\pm$0.34} & 68.42{\scriptsize$\pm$0.82} & 41.60 ($\uparrow$0.48) \\
& \cellcolor{blue!10}\textbf{\method{} (Ours)}
& \cellcolor{blue!10}\underline{18.27{\scriptsize$\pm$0.49}}
& \cellcolor{blue!10}\underline{49.89{\scriptsize$\pm$1.44}}
& \cellcolor{blue!10}\textbf{55.67{\scriptsize$\pm$0.76}}
& \cellcolor{blue!10}\textbf{29.01{\scriptsize$\pm$0.39}}
& \cellcolor{blue!10}\textbf{73.00{\scriptsize$\pm$0.67}}
& \cellcolor{blue!10}\textbf{45.17} ($\uparrow$4.04) \\
\midrule
\multirow{8}{*}{InternVL3.5-4B}
& Base  & 14.76{\scriptsize$\pm$0.69} & 49.13{\scriptsize$\pm$2.78} & 61.76{\scriptsize$\pm$0.99} & 33.60{\scriptsize$\pm$0.25} & 70.47{\scriptsize$\pm$0.77} & 45.95 \\
& GRPO  & 14.85{\scriptsize$\pm$0.56} & 51.09{\scriptsize$\pm$1.35} & 62.01{\scriptsize$\pm$1.34} & \underline{34.89{\scriptsize$\pm$0.45}} & 71.58{\scriptsize$\pm$1.94} & 46.88 ($\uparrow$0.93) \\
& PAPO  & 14.74{\scriptsize$\pm$0.38} & 49.69{\scriptsize$\pm$1.16} & 62.28{\scriptsize$\pm$1.15} & 34.67{\scriptsize$\pm$0.78} & 71.24{\scriptsize$\pm$0.37} & 46.53 ($\uparrow$0.58) \\
& VPPO  & \underline{15.68{\scriptsize$\pm$0.72}} & \underline{52.52{\scriptsize$\pm$1.71}} & \underline{62.36{\scriptsize$\pm$1.24}} & 34.46{\scriptsize$\pm$0.50} & \underline{72.03{\scriptsize$\pm$0.86}} & \underline{47.41} ($\uparrow$1.46) \\
& OPSD  & 14.53{\scriptsize$\pm$0.67} & 52.27{\scriptsize$\pm$1.49} & 61.03{\scriptsize$\pm$0.77} & 33.78{\scriptsize$\pm$0.65} & 71.03{\scriptsize$\pm$0.68} & 46.53 ($\uparrow$0.58) \\
& RLSD  & \textbf{15.92{\scriptsize$\pm$0.91}} & 50.56{\scriptsize$\pm$1.84} & 61.56{\scriptsize$\pm$0.82} & 34.66{\scriptsize$\pm$0.68} & 71.71{\scriptsize$\pm$0.56} & 46.88 ($\uparrow$0.93) \\
& RLCSD & 14.69{\scriptsize$\pm$0.40} & 49.22{\scriptsize$\pm$2.52} & 61.55{\scriptsize$\pm$0.73} & 33.75{\scriptsize$\pm$0.62} & 70.37{\scriptsize$\pm$0.73} & 45.92 ($\downarrow$0.03) \\
& \cellcolor{blue!10}\textbf{\method{} (Ours)}
& \cellcolor{blue!10}15.54{\scriptsize$\pm$0.59}
& \cellcolor{blue!10}\textbf{52.80{\scriptsize$\pm$2.58}}
& \cellcolor{blue!10}\textbf{63.78{\scriptsize$\pm$1.01}}
& \cellcolor{blue!10}\textbf{35.26{\scriptsize$\pm$0.37}}
& \cellcolor{blue!10}\textbf{72.60{\scriptsize$\pm$1.15}}
& \cellcolor{blue!10}\textbf{48.00} ($\uparrow$2.05) \\
\midrule
\multirow{8}{*}{\makecell{Qwen3-VL-2B\\Instruct}}
& Base  & 10.32{\scriptsize$\pm$0.53} & 33.61{\scriptsize$\pm$1.30} & 47.54{\scriptsize$\pm$0.77} & 20.37{\scriptsize$\pm$0.56} & 58.36{\scriptsize$\pm$0.57} & 34.04 \\
& GRPO  & 9.50{\scriptsize$\pm$0.61} & 42.86{\scriptsize$\pm$1.53} & 49.17{\scriptsize$\pm$0.92} & 22.77{\scriptsize$\pm$0.44} & 52.33{\scriptsize$\pm$1.16} & 35.32 ($\uparrow$1.28) \\
& PAPO  & 10.69{\scriptsize$\pm$0.28} & 39.09{\scriptsize$\pm$1.77} & 49.81{\scriptsize$\pm$1.11} & 24.80{\scriptsize$\pm$0.70} & 57.64{\scriptsize$\pm$0.48} & 36.41 ($\uparrow$2.37) \\
& VPPO  & 10.70{\scriptsize$\pm$0.29} & 39.99{\scriptsize$\pm$1.84} & \underline{50.96{\scriptsize$\pm$0.97}} & 23.34{\scriptsize$\pm$0.60} & \textbf{59.90{\scriptsize$\pm$0.67}} & 36.98 ($\uparrow$2.94) \\
& OPSD  & \underline{10.73{\scriptsize$\pm$0.51}} & 33.69{\scriptsize$\pm$1.63} & 47.98{\scriptsize$\pm$0.68} & 19.90{\scriptsize$\pm$0.66} & 58.09{\scriptsize$\pm$0.69} & 34.08 ($\uparrow$0.04) \\
& RLSD  & 10.71{\scriptsize$\pm$0.85} & \underline{44.42{\scriptsize$\pm$1.12}} & \textbf{51.71{\scriptsize$\pm$0.85}} & \textbf{28.90{\scriptsize$\pm$0.65}} & 58.16{\scriptsize$\pm$0.43} & \textbf{38.78} ($\uparrow$4.74) \\
& RLCSD & 10.62{\scriptsize$\pm$0.49} & 34.25{\scriptsize$\pm$0.62} & 47.74{\scriptsize$\pm$1.09} & 20.46{\scriptsize$\pm$0.42} & \underline{58.50{\scriptsize$\pm$0.64}} & 34.32 ($\uparrow$0.28) \\
& \cellcolor{blue!10}\textbf{\method{} (Ours)}
& \cellcolor{blue!10}\textbf{11.90{\scriptsize$\pm$0.51}}
& \cellcolor{blue!10}\textbf{46.98{\scriptsize$\pm$1.70}}
& \cellcolor{blue!10}50.56{\scriptsize$\pm$0.33}
& \cellcolor{blue!10}\underline{26.09{\scriptsize$\pm$0.26}}
& \cellcolor{blue!10}55.60{\scriptsize$\pm$0.79}
& \cellcolor{blue!10}\underline{38.23} ($\uparrow$4.19) \\
\bottomrule
\end{tabularx}
\caption{Accuracy (\%)  (mean@8) on five multimodal reasoning benchmarks with standard deviation.}
\label{tab:main_std}
\end{table*}

\subsection{Results on Qwen3-VL-2B-Instruct}
\label{app:qwen3vl2b}

On Qwen3-VL-2B-Instruct (Table~\ref{tab:main_std}), \method{} is second on average (38.23\%, +4.19 over Base) behind RLSD (38.78\%). It is best on GSM8K-V and LogicVista but trails RLSD on MMMU, MathVision, and MathVista, and falls below Base on MathVista (55.60 vs.\ 58.36), as GRPO does (52.33). We attribute this to the 2B model's limited capacity as a self-teacher. The training data---filtered to retain only problems that the much larger Qwen3-VL-4B-Thinking consistently fails---is too difficult for 2B to reliably solve even with privileged hints. When the 2B teacher itself produces low-confidence or inconsistent predictions under correct-answer contexts, the pairwise contrasts degrade into uninformative noise, and the confidence $S_i$ used by adaptive trajectory confidence becomes unreliable. Token-level credit assignment thus provides less benefit over uniform GRPO when the teacher capacity is insufficient relative to task difficulty.

\subsection{Full Ablation Results}
\label{app:ablation_full}

Table~\ref{tab:ablation_full} lists per-benchmark accuracy on Qwen3-VL-8B-Instruct for every ablation configuration, each appearing once. The full \method{} uses full $+$ null vision contrast with the $S_{\mathrm{mean}}$ confidence and $\lambda_i\in[0.15,0.65]$; other rows change only the listed setting. In terms of Eq.~\eqref{eq:token_contrast}, full vision, full $-$ null vision, and full $+$ null vision contrast correspond to $c=\Delta$, $c=2\Delta-\bar{\Delta}$, and $c=1.5\Delta+0.5\bar{\Delta}$, and the vision gap is $c=\ell^{+,V}-\ell^{+,\varnothing}$.

\begin{table*}[t]
\centering
\footnotesize
\setlength{\tabcolsep}{2.0pt}
\begin{tabularx}{\textwidth}{l *{6}{Y}}
\toprule
Variant & GSM8K-V & LogicVista & MMMU & MathVision & MathVista & Average \\
\midrule
\textbf{\method{}} & 28.13\scriptsize$\pm$0.52 & 63.31\scriptsize$\pm$1.52 & 69.00\scriptsize$\pm$0.87 & 54.01\scriptsize$\pm$0.42 & 77.90\scriptsize$\pm$0.83 & 58.47 \\
\midrule
\multicolumn{7}{@{}l}{\textbf{\textit{Token contrast, fixed $\lambda=0.5$}}} \\
Vision gap (no hint contrast) & 24.03\scriptsize$\pm$0.70 & 55.26\scriptsize$\pm$1.53 & 61.22\scriptsize$\pm$1.10 & 40.95\scriptsize$\pm$0.41 & 73.80\scriptsize$\pm$0.49 & 51.05 \\
Hint contrast, full vision & 26.91\scriptsize$\pm$0.70 & 61.97\scriptsize$\pm$1.33 & 65.56\scriptsize$\pm$1.09 & 53.36\scriptsize$\pm$0.71 & 77.50\scriptsize$\pm$0.57 & 57.06 \\
Hint contrast, full $-$ null vision & 28.28\scriptsize$\pm$0.58 & 60.63\scriptsize$\pm$1.44 & 67.33\scriptsize$\pm$1.07 & 53.36\scriptsize$\pm$0.47 & 78.80\scriptsize$\pm$0.62 & 57.68 \\
Hint contrast, full $+$ null vision & 26.23\scriptsize$\pm$0.85 & 61.07\scriptsize$\pm$1.33 & 66.67\scriptsize$\pm$0.83 & 53.06\scriptsize$\pm$0.73 & 79.10\scriptsize$\pm$0.90 & 57.23 \\
\midrule
\multicolumn{7}{@{}l}{\textbf{\textit{Fixed mixing rate, full $+$ null vision contrast}}} \\
$\lambda=0.25$ & 26.00\scriptsize$\pm$0.32 & 61.97\scriptsize$\pm$1.32 & 67.33\scriptsize$\pm$0.84 & 51.05\scriptsize$\pm$0.55 & 76.30\scriptsize$\pm$0.55 & 56.53 \\
$\lambda=0.40$ (adaptive median) & 27.52\scriptsize$\pm$0.62 & 61.52\scriptsize$\pm$1.12 & 67.00\scriptsize$\pm$0.99 & 52.14\scriptsize$\pm$0.50 & 78.30\scriptsize$\pm$0.73 & 57.30 \\
$\lambda=0.50$ & 26.23\scriptsize$\pm$0.52 & 61.07\scriptsize$\pm$1.05 & 66.67\scriptsize$\pm$0.85 & 53.06\scriptsize$\pm$0.39 & 79.10\scriptsize$\pm$0.56 & 57.23 \\
$\lambda=0.75$ & 26.61\scriptsize$\pm$0.59 & 66.00\scriptsize$\pm$1.37 & 65.89\scriptsize$\pm$0.81 & 52.43\scriptsize$\pm$0.79 & 77.20\scriptsize$\pm$0.64 & 57.63 \\
$\lambda=1.00$ & 27.14\scriptsize$\pm$0.36 & 63.53\scriptsize$\pm$1.55 & 65.78\scriptsize$\pm$0.67 & 51.74\scriptsize$\pm$0.53 & 77.50\scriptsize$\pm$0.46 & 57.14 \\
\midrule
\multicolumn{7}{@{}l}{\textbf{\textit{Adaptive trajectory confidence}}} \\
$S_{\mathrm{std}}$ (dispersion-based), $\lambda_i\in[0.25,0.65]$ & 27.37\scriptsize$\pm$0.56 & 62.42\scriptsize$\pm$1.44 & 67.33\scriptsize$\pm$1.11 & 52.86\scriptsize$\pm$0.72 & 78.10\scriptsize$\pm$0.46 & 57.62 \\
Mix of $S_{\mathrm{mean}}$ and $S_{\mathrm{std}}$, $\lambda_i\in[0.25,0.65]$ & 28.58\scriptsize$\pm$0.61 & 62.42\scriptsize$\pm$1.18 & 67.56\scriptsize$\pm$0.92 & 52.76\scriptsize$\pm$0.52 & 76.90\scriptsize$\pm$0.59 & 57.64 \\
$S_{\mathrm{mean}}$, $\lambda_i\in[0.25,0.65]$ & 25.70\scriptsize$\pm$0.71 & 64.43\scriptsize$\pm$1.04 & 67.89\scriptsize$\pm$0.93 & 51.97\scriptsize$\pm$0.36 & 78.00\scriptsize$\pm$0.64 & 57.60 \\
$S_{\mathrm{mean}}$, $\lambda_i\in[0.10,0.65]$ & 26.16\scriptsize$\pm$0.50 & 61.30\scriptsize$\pm$1.44 & 66.89\scriptsize$\pm$0.72 & 53.36\scriptsize$\pm$0.48 & 77.50\scriptsize$\pm$0.94 & 57.04 \\
$S_{\mathrm{mean}}$, $\lambda_i\in[0.15,0.75]$ & 27.22\scriptsize$\pm$0.67 & 61.52\scriptsize$\pm$1.69 & 67.67\scriptsize$\pm$0.67 & 52.60\scriptsize$\pm$0.41 & 78.10\scriptsize$\pm$0.74 & 57.42 \\
$S_{\mathrm{mean}}$ on full $-$ null vision contrast & 27.98\scriptsize$\pm$0.44 & 61.07\scriptsize$\pm$1.20 & 67.67\scriptsize$\pm$0.65 & 52.20\scriptsize$\pm$0.52 & 77.80\scriptsize$\pm$0.93 & 57.34 \\
\midrule
\multicolumn{7}{@{}l}{\textbf{\textit{Contrast coefficients $(\alpha,\beta)$, $S_{\mathrm{mean}}$ confidence}}} \\
$(\alpha,\beta)=(1.00,\,0.00)$ & 27.82\scriptsize$\pm$0.48 & 61.30\scriptsize$\pm$1.09 & 67.89\scriptsize$\pm$1.12 & 53.19\scriptsize$\pm$0.44 & 78.40\scriptsize$\pm$0.61 & 57.72 \\
$(\alpha,\beta)=(1.125,\,0.125)$ & 26.99\scriptsize$\pm$0.70 & 61.97\scriptsize$\pm$1.35 & 67.89\scriptsize$\pm$0.89 & 53.06\scriptsize$\pm$0.66 & 77.40\scriptsize$\pm$0.52 & 57.46 \\
$(\alpha,\beta)=(1.25,\,0.25)$ & 27.07\scriptsize$\pm$0.74 & 63.31\scriptsize$\pm$1.36 & 67.22\scriptsize$\pm$0.70 & 54.28\scriptsize$\pm$0.74 & 78.40\scriptsize$\pm$0.61 & 58.06 \\
$(\alpha,\beta)=(1.75,\,0.75)$ & 28.20\scriptsize$\pm$0.60 & 62.86\scriptsize$\pm$1.68 & 65.44\scriptsize$\pm$0.69 & 51.81\scriptsize$\pm$0.78 & 77.30\scriptsize$\pm$0.89 & 57.12 \\
$(\alpha,\beta)=(2.00,\,1.00)$ & 26.69\scriptsize$\pm$0.65 & 61.30\scriptsize$\pm$1.02 & 69.22\scriptsize$\pm$0.97 & 53.03\scriptsize$\pm$0.61 & 76.50\scriptsize$\pm$0.91 & 57.35 \\
\bottomrule
\end{tabularx}
\caption{Per-benchmark accuracy (\%) of all ablation configurations (mean@8) with standard deviation.}
\label{tab:ablation_full}
\end{table*}

\section{Case Study}
\label{app:case_study}

\begin{figure*}[t]
    \centering
    \includegraphics[width=0.92\textwidth]{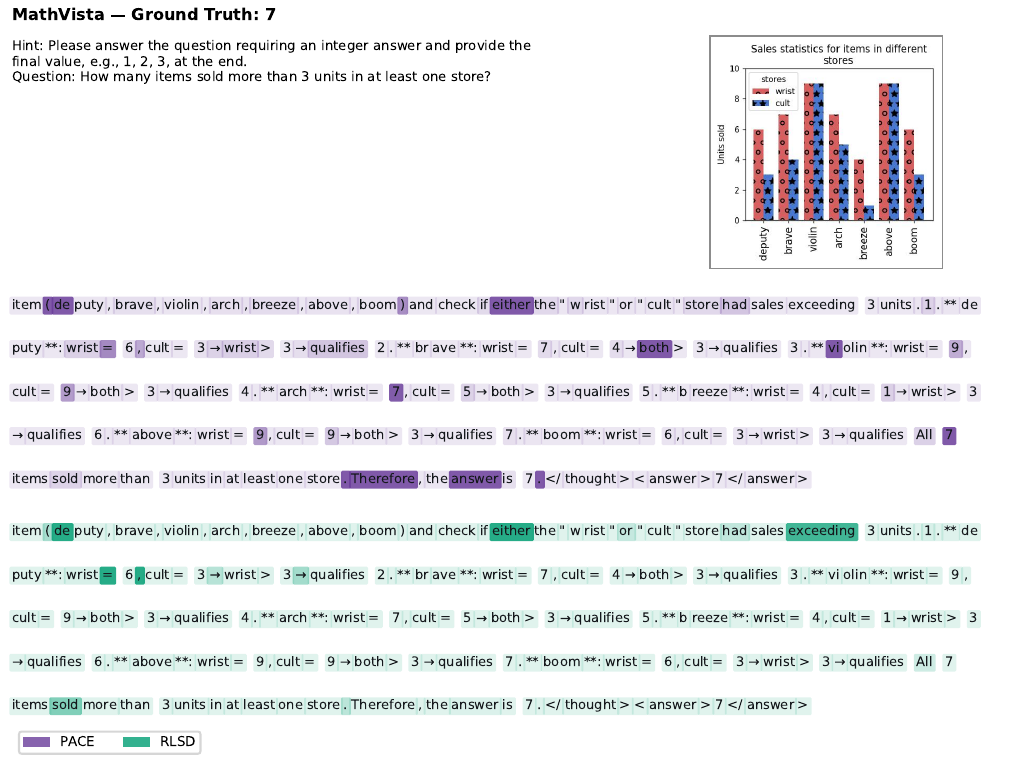}
    \caption{Case Study.}
    \label{fig:case_mathvista_count}
\end{figure*}

\begin{figure*}[t]
    \centering
    \includegraphics[width=0.92\textwidth]{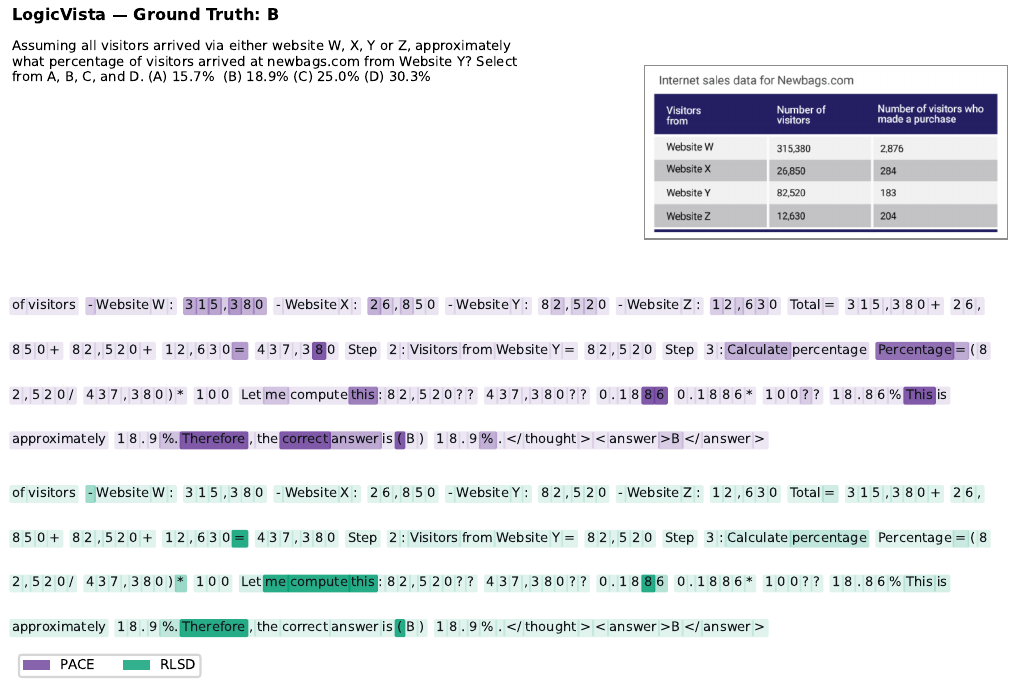}
    \caption{Case Study.}
    \label{fig:case_logicvista}
\end{figure*}

\begin{figure*}[t]
    \centering
    \includegraphics[width=0.92\textwidth]{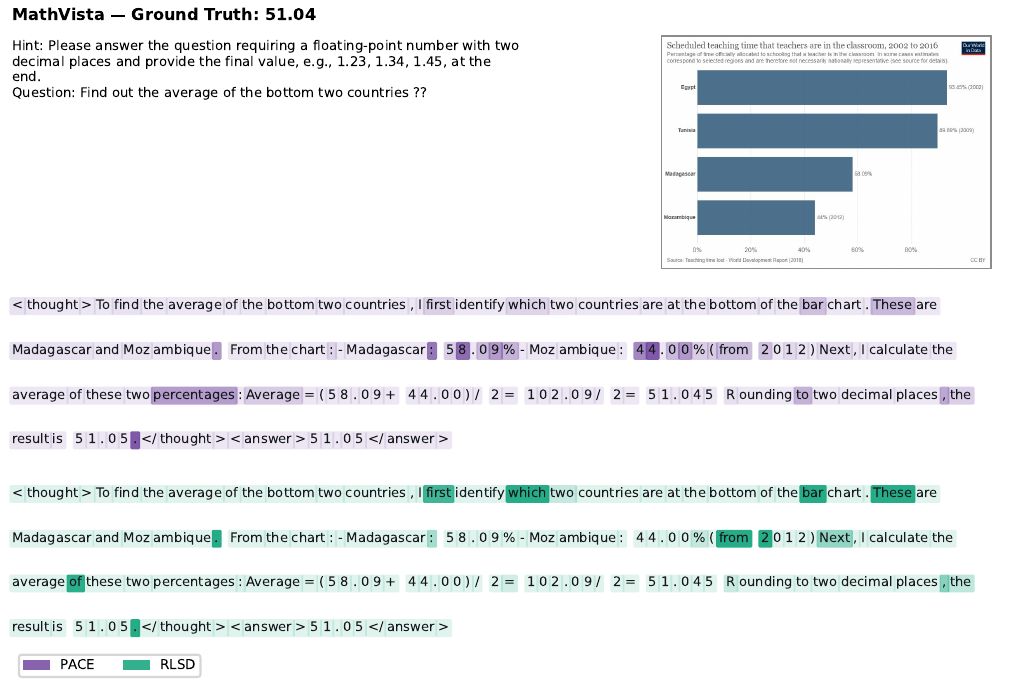}
    \caption{Case Study.}
    \label{fig:case_mathvista_avg}
\end{figure*}

Figures~\ref{fig:case_mathvista_count}--\ref{fig:case_mathvista_avg} visualize token credit on three responses from the evaluation benchmarks. For each response, the upper copy shows the token weights assigned by \method{} and the lower copy those assigned by RLSD on the same sampled tokens; darker shading marks a larger weight, i.e.\ a larger share of the trajectory advantage. Both methods keep the sign and mean of the trajectory advantage, so the figures differ only in \emph{where} credit is placed within the response.

\paragraph{Credit on visual readouts (Figure~\ref{fig:case_mathvista_count}).}
The MathVista question asks how many items sold more than 3 units in at least one store of a grouped bar chart. The response reads both bars for all seven items and concludes that all seven qualify. \method{} shades the values read from the chart (e.g., 9 for \emph{violin}, 7 for \emph{arch}, 1 for \emph{breeze}), the comparison token ``both'', and the final count ``All 7''. RLSD instead highlights tokens that restate the question (``either'', ``exceeding'', ``sold'') and leaves the chart readouts nearly unweighted. The contrast between correct and incorrect hints singles out the tokens that decide the count, rather than restating the question.

\paragraph{Credit on computation (Figure~\ref{fig:case_logicvista}).}
The LogicVista question asks for Website Y's share of visitors in a table. The response copies the four visitor counts, sums them to 437,380, and divides to obtain 18.86\%, which it rounds to option (B). \method{} places its largest weights on the result-bearing tokens: the copied counts (e.g., 315,380), the distinguishing digit of the total, the quotient digits of 0.1886, and the decision tokens ``Percentage ='', ``Therefore'', ``correct answer'', and ``(B''. RLSD concentrates on the connective phrase ``me compute this'' and on ``='' and ``Therefore'', and assigns nearly uniform weight to the copied counts.

\paragraph{Credit on an incorrect response (Figure~\ref{fig:case_mathvista_avg}).}
The third response is incorrect: it reads 58.09\% and 44.00\% for the bottom two countries and answers 51.05 instead of 51.04. Its arithmetic is consistent with these readouts, so the error lies in the extracted values or the rounding. Because the trajectory advantage is negative, a larger weight here means a larger share of the penalty. \method{} concentrates the penalty on the extracted values and their context (``Madagascar:'', the digits of 58.09 and 44.00, ``from 2012'') and on ``percentages'', while RLSD spreads it over planning tokens such as ``first'', ``which'', ``bar'', ``These'', and ``Next''. Concentrating the penalty on the readouts, rather than on generic planning steps, spares the parts of the reasoning that are correct.

These examples are qualitative and were selected for illustration, which show how the two signals differ in placement, not that \method{} assigns credit correctly on every response.

\clearpage

\end{document}